\documentclass[12pt]{article}

\usepackage{xcolor}
\usepackage[colorinlistoftodos,prependcaption,textsize=tiny]{todonotes}

\usepackage{newtxtext,newtxmath}

\usepackage{graphicx}

\usepackage[letterpaper,margin=1in]{geometry}

\usepackage{colortbl}
\usepackage{circuitikz}

\usepackage{chemformula}

\usepackage{tikz}
\usepackage{quantikz}
\usepackage{enumitem} 

\usepackage{booktabs} 
\usepackage{array}    
\usepackage{csvsimple}
\usepackage{siunitx}
\usepackage{longtable}
\usepackage{algorithm}
\usepackage{algorithmic}

\usepackage{subcaption} 

\usepackage{longtable}
\renewcommand{\arraystretch}{0.55}

\def\L{\mathcal{L}}
\def\cP{\mathcal{P}}
\def\Tr{\operatorname{Tr}}
\def\LSA{\operatorname{LSA}}

\renewenvironment{abstract}
	{\quotation}
	{\endquotation}

\date{}

\makeatletter
\renewcommand{\fnum@figure}{\textbf{Figure \thefigure}}
\renewcommand{\fnum@table}{\textbf{Table \thetable}}
\makeatother

\usepackage{scicite}

\usepackage{url}

\def\scititle{
	Discovering  Algorithms with\\ Computational Language Processing 
}

\title{\bfseries \boldmath \scititle}

\author{
	Th\'{e}o Bourdais$^{1}$,
	Abeynaya Gnanasekaran$^{2}$,
    Houman Owhadi$^{1}$,
    Tuhin Sahai$^{2*}$\and
	\small$^{1}$California Institute of Technology, Pasadena, CA 91125, USA.\and
	\small$^{2}$SRI International, Menlo Park, CA 94025, USA.\and
	\small$^\ast$Corresponding author. Email: tuhin.sahai@sri.com\and
	\small Authors are listed alphabetically and contributed equally to this work.
}

\begin{document} 

\maketitle

\begin{abstract} \bfseries \boldmath
Algorithms are the engine for reproducible problem-solving.
We present a framework  automating algorithm discovery by conceptualizing them as 
sequences of operations, represented as tokens. These computational tokens are chained using a  grammar, enabling the formation of increasingly sophisticated  procedures. Our ensemble Monte Carlo tree search (MCTS) guided by reinforcement learning (RL) explores token chaining and drives the creation of new tokens. This methodology rediscovers, improves, and generates new  algorithms  that substantially outperform  existing methods for strongly NP-hard combinatorial optimization  problems and foundational quantum computing approaches such as Grover's  and  Quantum Approximate Optimization Algorithm. Operating at the computational rather than code-generation level, our framework produces algorithms that can be tailored specifically to problem instances, not merely classes.
\end{abstract}
\begin{figure}
    \centering
\includegraphics[width=0.8\linewidth]{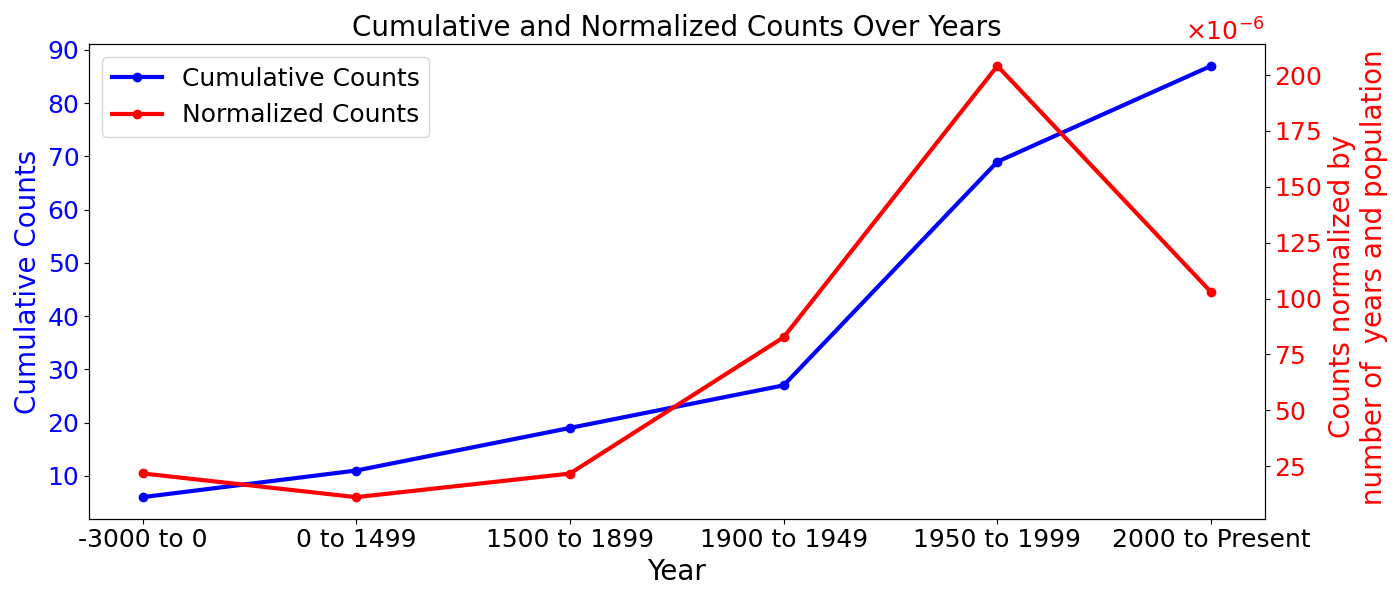}
    
    \caption{Historical timeline of key algorithmic breakthroughs (full list  in supplementary).}
    \label{figalgovstime}
\end{figure}

\noindent Although major algorithms continue to be introduced at a rapid pace (see Fig.~\ref{figalgovstime}), the rate of this growth appears to slow when adjusted for population size. Moreover, the discovery of new algorithms predominantly remains reliant upon trial-and-error methods, intuition, and informed guesswork.
While various strategies have attempted to systematize this discovery process, these approaches typically remain problem-specific, as exemplified by fast solver design \cite{OwhScobook2018} or the development of matrix multiplication algorithms \cite{fawzi2022discovering}. Alternatively, some approaches adopt direct code-generation paradigms such as AlphaEvolve \cite{novikov2025alphaevolve} in which ensembles of large language models (LLMs) serve as coding agents. These are often integrated with evaluation metrics and evolutionary strategies, such as genetic algorithms, to iteratively refine large code segments \cite{lehman2023evolution, romera2024mathematical, liu2024evolution}. AlphaEvolve and other existing methods produce singular solutions that are applied to every instance within a problem class. Therefore, (i) they are constrained by the no free lunch theorem~\cite{wolpert1997no}, that essentially states that no single algorithm can universally outperform all others across every possible problem instance; and (ii) algorithms that \emph{exploit} instance specific structure are key for the building performant solutions to important problems.
In this paper, we address this issue and introduce a significantly more informed (and thereby efficient) approach by first tokenizing  the computing process itself, an abstraction crucial for simplifying and generalizing complex algorithmic representations, which we term computational language processing (CLP). Next, we introduce a RL \emph{ensemble} variant of MCTS to efficiently explore the space of grammatically consistent token chains.
Contrary to the prevalent paradigm, where an algorithm is designed for an entire class of problems (e.g., the QAP) and uniformly applied across all instances, our framework is instance-adaptive and performs reinforcement learning directly at the computational level of each specific problem instance. As a result, our approach can generate both problem class algorithms and \emph{distinct, instance-specific algorithms finely tailored to the nuances of individual problems}.
We demonstrate the effectiveness of this approach in algorithmic discovery through applications to the Quadratic Assignment Problem (a generic strongly NP-hard combinatorial optimization problem to which many other NP-hard problems can be reduced, see supplementary text Sec.~\ref{Secqaphard}), the quantum (unstructured) search problem and Quantum Approximate Optimization Algorithm. 
Crucially, this high-level framework not only considerably \emph{simplifies the conceptualization and generalization of complex algorithms} and integrates naturally with downstream code-generation methods, but it also generalizes naturally to broader domains, including systems engineering (the design of complex physical systems/processes) and mathematical discovery.

\begin{figure}[h]
    \centering
    \includegraphics[width=1\linewidth]{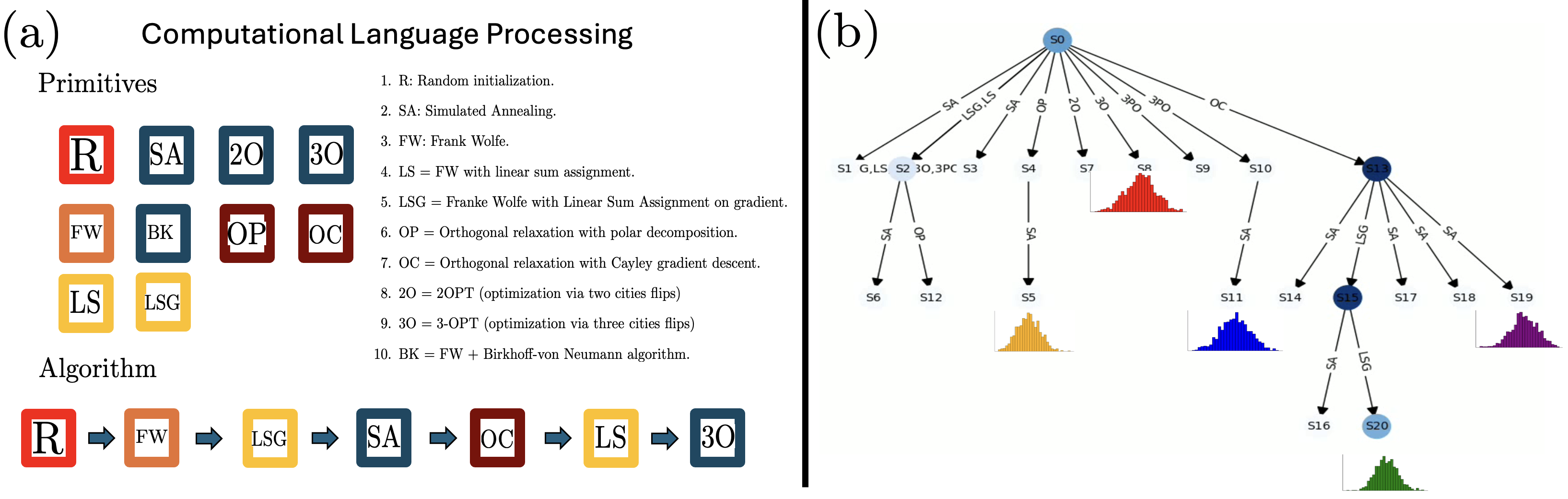}
    \caption{(a) We conceptualize an algorithm as a chain of elementary computational steps (primitives/tokens), chained with a precise grammar/syntax. (b) Starting with elementary tokens, we discover new tokens and algorithms with ensemble
MCTS + RL as a babbling mechanism
to search over the tree of all possible strings of tokens. Edges represent primitives/tokens and nodes  the ensemble of computational state variables sampled from the string of (possibly stochastic) tokens leading that node. 
}
    \label{figscience2}
\end{figure}

\begin{figure}[h]
    \centering
    \includegraphics[width=1\linewidth]{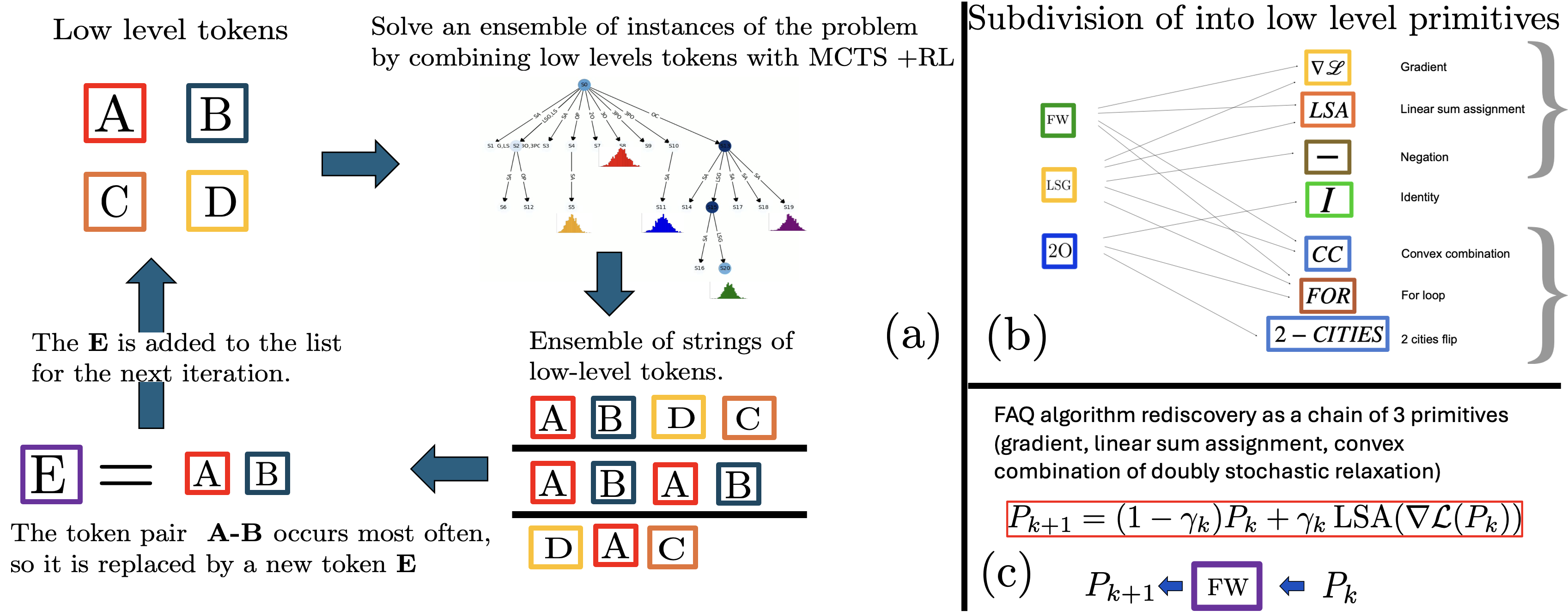}
    \caption{(a) Algorithmic Byte-Pair encoding to increase vocabulary by creating new tokens from old ones (b) Low-level primitives (c) Fast Approximate QAP (FAQ) algorithm rediscovery as a chain of 3 primitives (gradient, linear sum assignment, convex combination of doubly stochastic relaxation).
}
    \label{figscience3}
\end{figure}

\begin{figure}[h]
    \centering
    \includegraphics[width=\linewidth]{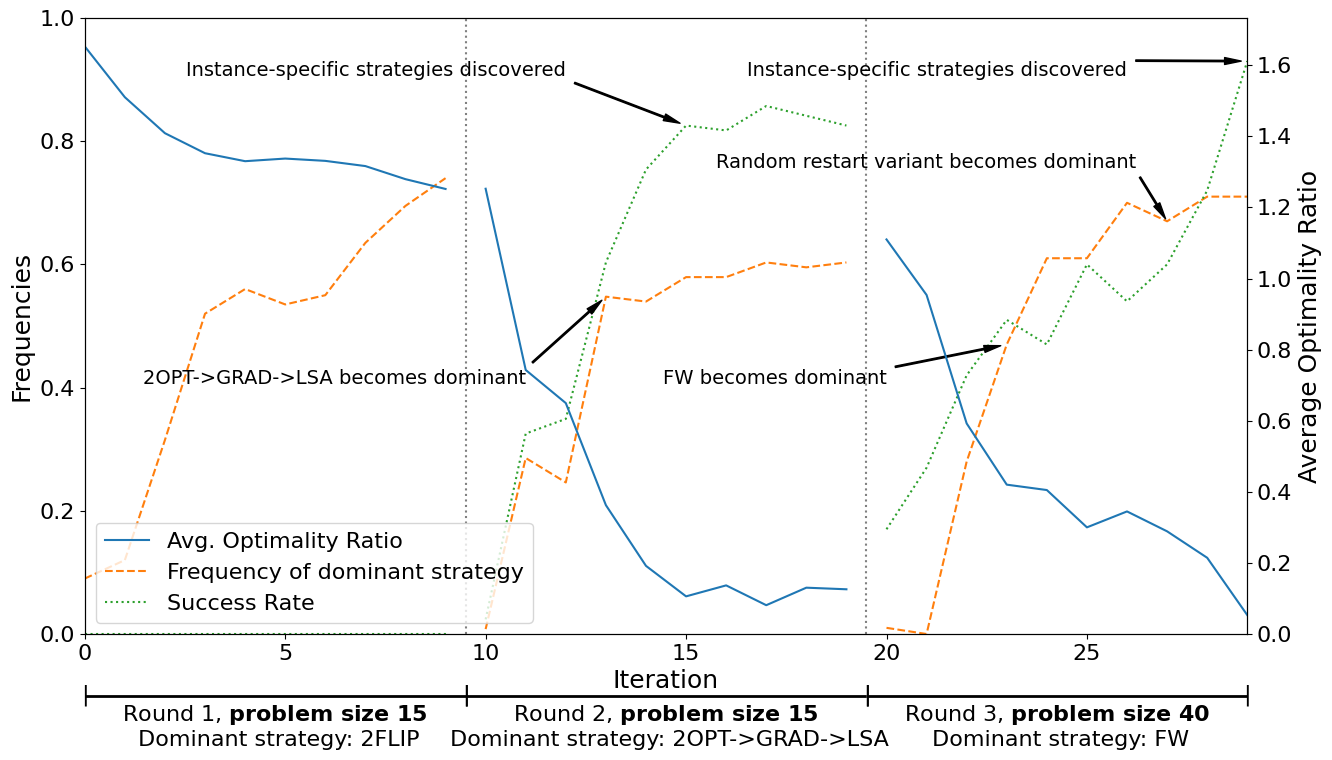}
    \caption{Emergence of dominant strategies as difficulty increases: (left-axis) Frequencies of occurrence of dominant strategy and finding the global optimum. (right-axis) Average relative optimality gap, $(\mathcal{L}-\min\mathcal{L})/\min\mathcal{L}$. (x-axis) Iterations, with a round of BPE every 10 iterations is marked by a vertical dotted line. CLP discovers increasingly more complex and effective algorithms, with identifiable dominant strategies and instance specific adaptations, to reach a high success rate. Details in \ref{secs15}}
    \label{fig:low_level_primitives}
\end{figure}

\paragraph{Computational Language Processing.}
Our framework conceptualizes  an algorithm as 
``a sequence of finite computational steps that transform an input into a desired output'' \cite{cormen2009introduction}. Since these computational steps naturally correspond to computational graphs 
\cite{owhadi2022computational,bourdais2024codiscovering}, the problem of discovering an algorithm reduces to  discovering a computational graph whose structure can be represented as a sequence of known elementary computational graphs/steps which we call \emph{primitives}.
Drawing a parallel between computational primitives and letters, we  view an algorithm as a sentence (Fig.~\ref{figscience2}(a)), and algorithm discovery as learning to speak in a computational language. This perspective motivates a language-based learning approach, abstracting away from domain-specific tasks to focus on chaining arbitrary computational steps. This enables algorithm discovery across diverse problem domains with a \emph{unified  methodology}.
Therefore, analogous to natural language, the set of primitives serves as the  \emph{the alphabet} for our computational language. The discovery of turbo codes (which have transformed digital communications by enabling efficient and reliable data transmission close to the theoretical maximum) illustrates the idea of composing primitives to generate algorithms precisely.  
Indeed, as reflected by Claude Berrou and Alain Glavieux, turbo codes emerged as ``the result of an empirical, painstaking construction of a global coding/decoding scheme, using existing bricks that had never been put together in this way before'' \cite{Guizzo2004TurboCodes}.
Unlike natural language, no large dataset of training examples exist for our task. To address this our framework alternates between two complementary phases (see Fig.~\ref{figscience2}(a),~\ref{figscience3}(a), and~\ref{fig:low_level_primitives}), (i) an ensemble MCTS and RL paradigm, to automate and scale the exploration of chains of computational steps. It composes elementary primitives into progressively more complex operations (analogous to forming words in natural language) which we term \emph{tokens}.
(ii) Incrementally expanding our computational vocabulary by merging the most efficient strings of tokens, inspired by methods such as Byte-Pair Encoding (BPE)~\cite{gage1994new}.

\paragraph{Ensemble Monte Carlo tree search with reinforcement learning.}
To  explore and refine the chaining of computational primitives, we integrate MCTS~\cite{browne2012survey} with trained policy and value neural networks. MCTS incrementally builds search trees using random sampling and targeted exploration, efficiently focusing on promising moves without exhaustively searching the entire tree. Combined with deep neural networks and self-play, this strategy reached breakthroughs like AlphaGo's superhuman performance in Go and AlphaZero's rapid dominance in chess \cite{silver2017mastering, silver2018general, schrittwieser2020mastering}. Our approach is a new ensemble variant of the AlphaZero reinforcement learning framework as detailed in Sec.~\ref{SecMCTSsupp} of the supplementary information. 
In this variant, edges correspond to computational actions represented by tokens, and each node maintains an ensemble of computational states derived from the sequence of tokens leading to that node (Fig.~\ref{figscience2}(b)). The framework is specifically designed to handle stochastic operations such as random initializations (e.g., random permutations) or stochastic computational operations, as they are key building blocks for designing efficient algorithms. In addition to token sequences, Boltzmann-weighted averages (akin to a softmax) of computational states are computed and serve as input features for the policy and value neural networks.
Tailored to algorithm discovery, our approach learns to chain tokens into algorithms \emph{dynamically tailored to specific problem instances}. Furthermore, when a single algorithm proves effective across all instances of a problem class, the learned policy tends to converge toward that solution. This policy convergence enables the discovery of general-purpose algorithms applicable to entire classes of problems when possible.

\paragraph{Algorithmic byte-pair encoding and  learning  higher-complexity tokens.}

The choice of primitives, i.e. the computational alphabet in CLP, directly governs the trade-off between expressivity and complexity of the search space. Imposing strong prior assumptions on the structure or choice of primitives introduces a significant bias towards the form of algorithm that is ultimately discovered. Conversely, while low-level primitives such as arithmetic operations or machine code offer maximum granularity, they also dramatically enlarge the search space, making discovery intractable. As a result,  coding agents based on LLMs rely heavily on supervised pretraining to serve as frameworks for algorithm discovery. This inherently biases them towards existing algorithms and  known patterns. Even when fine-tuned using RL, the often-used KL penalty limits exploration by anchoring updates to the pretrained distribution. This results in distributional shifts rather than genuine capability expansion \cite{korbak2022rl, vassoyan2025ignore}. Additionaly, the vast parameter count of LLMs makes RL-based exploration at inference computationally prohibitive. In contrast, we define a discrete set of computational tokens, allowing us to precisely control the amount of prior structure and eliminate the need for pretraining. This enables unbounded exploration driven solely by RL, while drastically reducing model size. \\
However, a fixed, manually designed vocabulary may constrain expressivity and necessitate long token sequences to represent complex behaviors, thereby increasing the search space and computational burden. To address this, we draw inspiration from NLP again to automatically discover tokens most adapted to the task at hand. Since we view algorithm discovery as learning to speak a computational language, RL rollouts can be viewed as ``babbling'', i.e. learning speech through trial and error. We collect these rollouts to construct a corpus of effective computational sequences for the current problem. Using this corpus, we apply a variant of the BPE algorithm, termed Algorithmic Byte-Pair Encoding (A-BPE), to iteratively merge frequently co-occurring token sequences into new, higher-level tokens. These tokens represent higher-level computational steps and can themselves be recursively composed, supporting hierarchical abstraction. This enables the iterative construction of complex algorithms and a dynamic combination of low and high-level representations (see Fig.\ref{figscience3}(a),
Fig.~\ref{fig:low_level_primitives}, and Sec.\ref{secs14}–\ref{secs15}).

\paragraph{The Quadratic Assignment Problem (QAP).}
The QAP is a particularly important combinatorial optimization problem that arises in a wide variety of settings. Often regarded as one of the most challenging NP-hard problems \cite{burkard1997qaplib}, that appears in supply chain optimization, where it is used to strategically place facilities in order to minimize item transit time or maximize supply chain throughput. Other applications include  airport design, data center optimization, and very large-scale integrated circuit design. For further discussion on the QAP and its importance, see  supplementary text Sec.~\ref{Secqapimportant}). Due to its combinatorial search space, and strongly NP-hard complexity 
(constant-factor approximations are NP-hard)
even moderate sized problems (e.g. size $n=20$) can be computationally challenging. 
Moreover, the QAP generalizes several classical optimization problems, for instance, setting $F_{i,j}=1$, symmetric distances with $D_{i,j}=D_{j,i}\geq 0$ and $D_{i,i}=0$, and $C_{i,j}=0$ in the QAP loss (Eqn.~\ref{eqqap1}) simplifies the QAP to the Traveling Salesman Problem (TSP). Similarly, setting $C=0$ and $D_{i,j}\in\{0,1\}$ reduces the QAP to the Graph Matching Problem \cite{sahai2019continuous}. Developing an automated framework that generates instance-adapted algorithms that outperform state-of-the-art solvers would, therefore, have significant industrial and scientific impact. 
To formally define the QAP, let $D$, $F$, and $C$ be $n \times n$ matrices with arbitrary real-valued entries, where $n \geq 2$. 
Let $\Pi_n$ be the set of all permutations of $\{1, 2, \ldots, n\}$,  the objective of the QAP, is to minimize $\min_{\pi \in \Pi_n} \L(\pi)$ where,
\begin{equation}\label{eqqap1}
\L(\pi):= \sum_{i=1}^n \sum_{j=1}^n F_{i,j} D_{\pi(i), \pi(j)} + \sum_{i=1}^n C_{i, \pi(i)}\, = \Tr\big[F P D^T P^T + C P^T\big]\,,
\end{equation}
and $P \in \cP_n$ is a $n \times n$ permutation matrix associated with $\pi$.

\paragraph{Primitives, tokens and Algorithmic Byte-Pair Encoding for the QAP.}
To illustrate the concepts of primitives and tokens, we first describe selected primitives commonly employed for solving the QAP, into which most existing algorithms naturally decompose (Fig.~\ref{figscience3}(a)--(c)). A representative subset of these primitives operates on a relaxation of the QAP from permutation matrices to the larger set of $n \times n$ \emph{doubly stochastic matrices}, denoted by $\tilde{\mathcal{P}}_n$. A matrix $P \in \tilde{\mathcal{P}}_n$ satisfies $P_{i,j} \in [0,1]$ for all entries, with row and column sums equal to one, i.e., $\sum_j P_{i,j}=1$ and $\sum_i P_{i,j}=1$.
By extending the loss function $\mathcal{L}$ to this relaxed domain, we obtain its gradient with respect to the matrix $P$ as follows:
$\nabla \mathcal{L}(P)= F P D^\top + F^\top P D + C$.
We introduce another fundamental primitive, \emph{interpolation between two doubly stochastic matrices}, defined by: $
P \;\leftarrow\; (1 - \gamma) \,Q \;+\; \gamma \,S$,
where $Q,S \in \tilde{\mathcal{P}}_n$ and $\gamma \in [0,1]$, resulting in another doubly stochastic matrix.
Additionally, given an arbitrary cost matrix $C \in \mathbb{R}^{n \times n}$, we consider the Linear Sum Assignment (LSA) problem, which seeks a permutation $\pi$ minimizing the total assignment cost $\sum_{i=1}^{n} C_{i,\pi(i)}$. The Hungarian (or Kuhn–Munkres) algorithm~\cite{kuhn1955hungarian, munkres1957algorithms} solves this problem optimally in $O(n^3)$ time. We denote the optimal permutation matrix solving the LSA defined by $C$ as:
$\LSA(C) := \text{argmin}_{P\in \mathcal{P}_n }\quad \sum_{i,j=1}^n C_{i,j} P_{i,j}$. Although we use the LSA algorithm as a primitive, in this case, we show how these algorithms can themselves be rediscovered using low-level primitives (see Fig.~\ref{fig:low_level_primitives}).
Additional primitives are detailed in Section~\ref{sec:all_primitives} of the supplementary information.
To illustrate vocabulary generation and token formation, consider the Frank–Wolfe algorithm's update step for the QAP:
\[
P_{k+1} = (1 - \gamma_k) P_k + \gamma_k\, \LSA(\nabla \mathcal{L}(P_k)).
\]
This update naturally decomposes into an interpolation token and a token representing the LSA operation applied to the gradient, each corresponding to previously defined computational primitives (Fig.~\ref{figscience3}(b,c)). Such decompositions demonstrate the incremental creation and complexity building inherent in our computational language framework.
Figure~\ref{fig:low_level_primitives} further demonstrates the practical application of our A-BPE strategy (see Fig.~\ref{figscience3}(a)) for token construction in the context of the QAP. Starting from a modest initial set of 8 elementary primitives (see Sec.~\ref{secs15}), the CLP approach leveraging A-BPE progressively discovers increasingly sophisticated and effective algorithms (tokens), emerging as dominant (most frequently employed) strategies (i.e., strings of tokens) over multiple rounds. For example, the Frank-Wolfe (FW) algorithm emerges as a dominant string of computational tokens. The resulting tokens exhibit both a recognizable dominant strategy and instance-specific adaptations, ultimately achieving a high success rates  (see Sec.~\ref{secs15} for details).

\begin{figure}[h]
    \centering
    \includegraphics[width=0.9\linewidth]{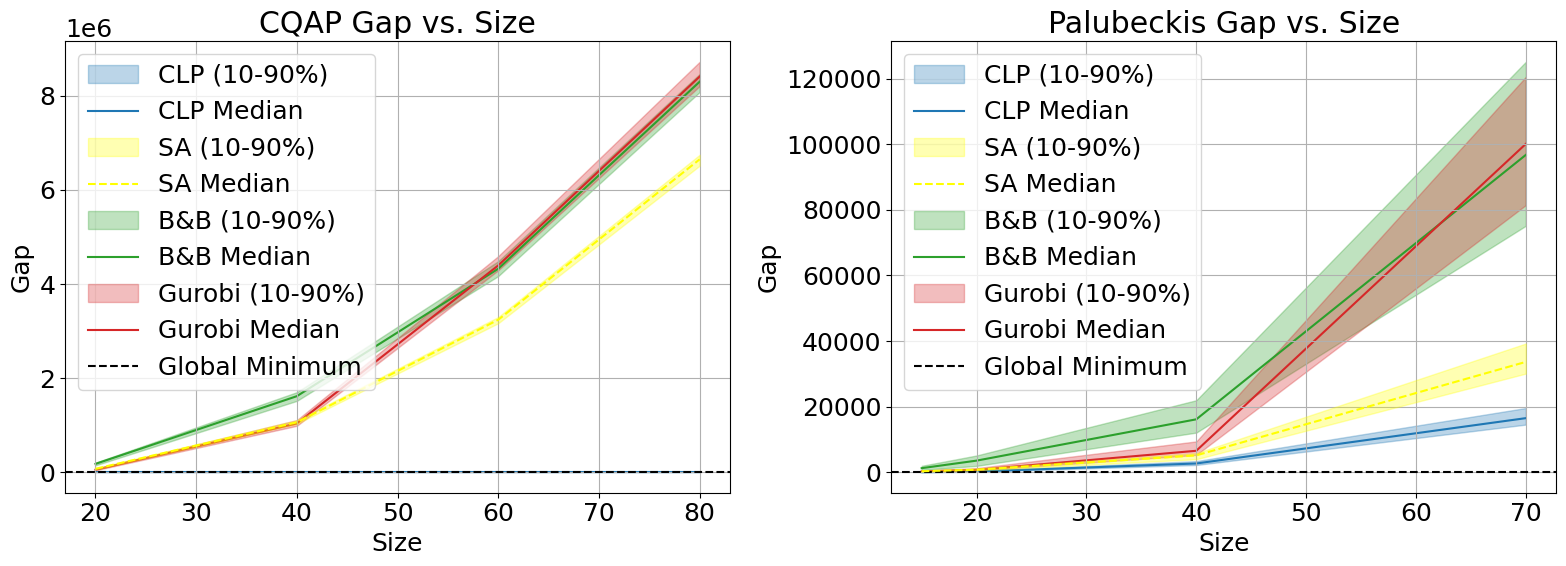}
    \caption{Optimality gap for different methods, with median, 10th and 90th quantile. On CQAP, CLP achieves global optimum on all tested instances.}
    \label{fig:qap-comparison}
\end{figure}

\paragraph{Results on the QAP.}
Algorithm discovery can be tackled at different levels of problem complexity, from hyperparameter tuning to heuristics design. To explore this spectrum, we evaluate our method on two benchmark generators, the \emph{composite} QAP (CQAP) \cite{drugan2015generating} and the significantly more challenging Palubeckis QAP (PQAP) \cite{palubeckis2000algorithm}, as well as the QAPLIB library of challenge problems. In three different experiments, CLP is provided with a base set of tokens and learns to compose them into increasingly sophisticated algorithms using a lightweight Transformer model guided by an ensemble variant of MCTS. With this approach, we obtain the following results: (i) Using well established algorithms as computational tokens, such as FW or 2-OPT (see Sec.~\ref{sec:all_primitives} for the entire list of tokens), our algorithm learns to chain these strategies to beat state-of-the-art meta-heuristics and the mixed-integer solver Gurobi. This performance is achieved across a range of problem sizes and instances with varying underlying structure, while keeping the number of objective evaluations constant. As shown in figure~\ref{fig:qap-comparison}, the resulting policies achieve optimal solutions in 100\% of tested instances from CQAP for $n\le80$, $\le1\%$ optimality gap on PQAP for $n\le80$ and beats or equals all other tested methods in 389 out of 390 tested PQAP and CQAP instances. On previously unseen QAPLIB examples, CLP generalizes to new problem structures, solving 41\% of QAPLIB instances to optimality and beating or equaling all other tested methods in 94.8\% of QAPLIB instances. See \ref{sec:high_level_primitives} for experimental details.
(ii) Starting from a minimal set of primitive operations such as identity operations, gradient steps, negation, Hungarian assignment (to map continuous matrices to permutation matrices), and simple control-flow tokens, CLP expands the library to 100+ composite tokens using A-BPE \cite{gage1994new}. After just three refinement rounds (of A-BPE and neural network retraining) it rediscovers many high-level tokens including $k$-OPT and FW. It also learns to perform random restarts for FW (see Fig.~\ref{figcycling}), enabling it to reach optimality in over 90\% of cases for CQAP instances of size 40, surpassing what standard implementations of FW achieve. See section \ref{secs15}.
(iii) When focusing on step-size optimization, CLP discovers a new cyclic step-size schedule that significantly outperforms the standard step-size schedule implemented within the \texttt{scipy} software package \cite{vogelstein2015fast} (see Fig.~\ref{figcycling}; see section \ref{sec:cycliclr}).

In short, CLP demonstrates effectiveness across all levels of algorithmic design: improving performance at a high-level, rediscovering known algorithms augmented with novel low-level restarting strategies, and optimizing associated parameters. CLP outperforms all baselines—including commercial solvers such as Gurobi—while scaling quadratically and providing fast answers on large problem instances. 

\paragraph{The Quantum Search Problem.}
Consider a finite, unstructured set $\mathcal{X}$ (with no inherent ordering or other distinguishing characteristics) containing $N$ elements, indexed by the set $\{0, 1, \dots, N-1\}$. Among the elements of $\mathcal{X}$, there is exactly one \textit{marked} or \textit{target} element $x^* \in \mathcal{X}$.
The task is to identify the marked element $x^*$ by querying an oracle (black-box) function $f: \mathcal{X} \to \{0,1\}$, where $f(x) = 1$ if  $x = x^*$ and $f(x) = 0$ otherwise.
 This oracle provides no additional information about the location of the target beyond the binary responses to queries.
The objective is to determine the marked element $x^*$ efficiently, minimizing the number of queries to the oracle function $f$.
In the quantum setting, the oracle is represented by a unitary operator $U_f$, acting on computational basis states $|x\rangle$ as
$U_f |x\rangle = (-1)^{f(x)}|x\rangle$. This oracle operator  flips the sign of the amplitude corresponding to the marked state.
Classically, since the set $\mathcal{X}$ is unstructured, the optimal solution is an exhaustive linear search whose complexity is $\mathcal{O}(N)$. Using the Quantum Oracle Representation, 
quantum computing leverages principles of superposition and amplitude amplification  to achieve a provable quantum advantage. Grover's algorithm~\cite{grover1996fast} accomplishes this, solving the quantum search problem in approximately
$\frac{\pi}{4}\sqrt{N}$ oracle queries with a high probability of success.  The quantum search problem is a cornerstone of quantum algorithmic theory because it: (a) Clearly demonstrates a quantum computational speed-up.
(b)  Has significant implications for cryptography, specifically affecting the computational security of cryptographic primitives reliant on brute-force resistance.
(c) Serves as an essential subroutine in quantum algorithms addressing optimization, combinatorial problems, and numerous other computational challenges.

\paragraph{Results on the Quantum Search Problem.}
An $n$-qubit quantum state $|\psi\rangle$ lives in a $2^n$-dimensional Hilbert space, and is expressed in the computational basis as:
$
|\psi\rangle = \sum_{x \in \{0,1\}^n} a_x |x\rangle$,  where $\sum_{x} |a_x|^2 = 1
$. Given a problem of size $N$ on $n=\log_2(N)$ qubits, we formulate the discovery of an algorithm for the quantum search problem as finding a sequence of quantum gates that maximizes the probability of measuring the target state with \emph{minimum circuit depth}. The CLP framework is setup with the following primitives (quantum gates) that make up Grover's algorithm: (I) The  Hadamard gate  on $n$ qubits whose action is on $x \in \{0,1\}^n$ is given by $H^{\otimes n}\vert x \rangle =  \frac{1}{\sqrt{2^n}}\sum_{y\in \{0,1\}^n}(-1)^{x\cdot y}\vert y\rangle$
where
$
x\cdot y = x_1 y_1 + x_2y_2 + \dots + x_ny_n \pmod{2}.
$
(II) The oracle  that marks the target state as $U_f|x\rangle = -|x\rangle$  if  $x = w$ and 
$U_f|x\rangle = |x\rangle$  if  $x \not= w$
where $|w\rangle$ is the target state. (III)  The Pauli-X (NOT) gate on $n$ qubits $
X^{\otimes n}|x_1 x_2 \dots x_n\rangle = |(1 - x_1)(1 - x_2)\dots(1 - x_n)\rangle$. (IV) The multi-controlled Z gate that applies a phase flip when all qubits are $|1\rangle$, i.e., 
$\operatorname{MCZ} |x_1 x_2 \dots x_n\rangle =
- |x_1 x_2 \dots x_n\rangle$ if  $x_1 = x_2 = \dots = x_n = 1$ and $\operatorname{MCZ} |x_1 x_2 \dots x_n\rangle =
|x_1 x_2 \dots x_n\rangle$ otherwise.
\begin{figure}[htbp]
    \centering
    \resizebox{\textwidth}{!}{%
    \begin{tabular}{c|c}
    \begin{quantikz} [wire types = {q, q, n, q, q}, transparent]
\lstick{$\ket{0}$} & \gate{H} & \gate[5]{U_f}  & \gate{H} \gategroup[5,steps=5,style={dashed,rounded corners,fill=blue!20, inner xsep=2pt},background,label style={label position=below,anchor=north,yshift=-0.2cm}]{{\sc $U_D$}} & \gate{X} & \ctrl{1} & \gate{X} & \gate{H} & \gate[5]{U_f} & \gate[5]{U_D} &  &  & \gate[5]{U_f} & \gate[5]{U_D}&\meter{}\\
\lstick{$\ket{0}$} & \gate{H} &              & \gate{H} & \gate{X} & \ctrl{0} & \gate{X} & \gate{H} &              &  &  &  &  & & \meter{}\\
&   \vdots   &  &\vdots & \vdots & \vdots & \vdots & \vdots  &  & & \cdots & \cdots & & & \vdots\\
 \lstick{$\ket{0}$} & \gate{H} &              & \gate{H} & \gate{X} & \ctrl{1} & \gate{X} & \gate{H} &              &  &  &  &  &  & \meter{}\\
\lstick{$\ket{0}$} & \gate{H} &              & \gate{H} & \gate{X} & \gate{Z} & \gate{X} & \gate{H} &              &  & & & & & \meter{}
\end{quantikz} 
        &
        \begin{quantikz}[wire types = {q, q, n, q, q}, transparent]
\lstick{$\ket{0}$} & \gate{X} & \gate{H} & \gate[5]{U_f} & \gate{H} \gategroup[5,steps=3,style={dashed,rounded corners,fill=blue!20, inner xsep=2pt},background,label style={label position=below,anchor=north,yshift=-0.2cm}]{{\sc $U'_D$}}& \ctrl{1} & \gate{H} & \gate[5]{U_f} &  \gate[5]{U'_D} &  &  & \gate[5]{U_f} &  \gate[5]{U'_D}& \meter{}\\
\lstick{$\ket{0}$} & \gate{X} & \gate{H} & & \gate{H} & \ctrl{0} & \gate{H} & &  & & & &&\meter{}\\
& \vdots &     \vdots  & & \vdots & \vdots & \vdots &   &  & \cdots & \cdots & &&\vdots\\
 \lstick{$\ket{0}$} & \gate{X} & \gate{H}   &    & \gate{H} & \ctrl{1} & \gate{H} &  &   &  &  & &&\meter{}\\
\lstick{$\ket{0}$} & \gate{X} & \gate{H} & & \gate{H} & \gate{Z} & \gate{H} & &  & & & && \meter{}
\end{quantikz}
    \end{tabular}}
    \caption{Comparison of Grover's algorithm implementations: standard (left, $U_D$ is the standard diffusion operator) and optimized (right, with simplified diffusion operator $U'_D$).}
    \label{fig:grover_comparison}
\end{figure}
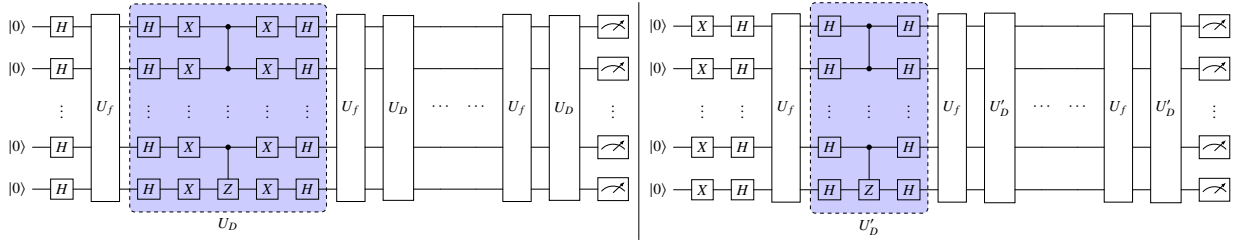

The  circuit discovered by our framework is shown in Fig.~\ref{fig:grover_comparison}. Compared to the standard Grover's  circuit, also shown in Figure~\ref{fig:grover_comparison}, \emph{the circuit depth is nearly halved}, yielding a  reduction of  $(\frac{\pi}{2}\sqrt{N}-1)$  layers, and the  number of gates is reduced by  $(\frac{\pi}{2}\sqrt{N}-1)\log N$. Quantum errors scale
exponentially with the number of qubits due to decoherence and noise. Consequently, our approach results in an exponential improvement in robustness measures.
The \emph{standard Grover's algorithm} can be summarized as follows. (1) Start with $|0\rangle^{\otimes n}$. Apply Hadamard gates to create uniform superposition:
      $ |s\rangle = H^{\otimes n}|0\rangle^{\otimes n} = \frac{1}{\sqrt{N}}\sum_{x=0}^{N-1}|x\rangle$. (2)  Repeat $O(\sqrt{N})$ times: (a) Apply oracle $U_f$ to mark the target state. (b) Apply diffusion operator $D = |s\rangle\langle s| - I$.
The \emph{optimized Grover's algorithm} can be summarized as follows. (1) Start with $|0\rangle^{\otimes n}$. (2) Apply X gates to all qubits: $X^{\otimes n}|0\rangle^{\otimes n} = |1\rangle^{\otimes n}$.  Apply Hadamard gates to create a different superposition:
   $    |s'\rangle = H^{\otimes n}|1\rangle^{\otimes n} =  \frac{1}{\sqrt{N}}\sum_{x=0}^{N-1}(-1)^{|x|}|x\rangle$
       where $|x|$ is the Hamming weight (number of 1's) in the binary representation of $x$. (3) Repeat $O(\sqrt{N})$ times: (a) Apply oracle $U_f$ to mark the target state. (b) Apply modified diffusion operator $D' = 2|s'\rangle\langle s'| - I$.
 We show in Sec.~\ref{SecsuppGrover} of the supplementary information that our discovered gate sequence is mathematically equivalent to Grover's algorithm and is an optimized version constructed using the same gate set. 
The key insight is that by starting with $|1\rangle^{\otimes n}$ instead of $|0\rangle^{\otimes n}$, we can directly implement the diffusion operator using just $H^{\otimes n} \rightarrow \text{MCZ}\rightarrow H^{\otimes n}$, eliminating the need for X gates at every iteration.

\paragraph{Results on the Quantum Approximate Optimization Algorithm (QAOA).} 
QAOA \cite{farhi2014quantum} is a hybrid quantum-classical optimization algorithm designed to approximately solve combinatorial optimization problems. In  Sec.~\ref{sec:clpqaoa}
we show that our approach achieves a 35\% improvement when compared with ADAPT-QAOA~\cite{zhu2022adaptive}, the adaptive version of QAOA  implemented within Nvidia's quantum computing library (CUDA-Q).

 \paragraph{Conclusion.}
In this work, we have demonstrated the versatility of CLP through three distinct examples for algorithm development. Our approach is broadly applicable beyond algorithmic development, offering significant potential for automating engineering system design by tokenizing complex designs into structured representations. Moreover, the capability of our framework to facilitate real-time algorithm development could enable emergent behaviors in agent-based systems, allowing adaptive and dynamic control in real-world scenarios.

\clearpage 
\bibliography{RPS, science_agis} 
\bibliographystyle{sciencemag}

%
%
%
%
%
%


\section*{Acknowledgments}
%
The authors thank Prof. Yannis Kevrekidis and Dr. Robert Lazar for insightful feedback and comments. 
This research was to a large extent initiated by Prof. Kevrekidis' vision articulated in his DARPA DIAL program (Mathematics for the Discovery of Algorithms and Architectures) whose objectives were to ``explore disruptive capabilities in computer-aided algorithm discovery via
optimization''.

\paragraph*{Funding:}
 This material is based upon work supported by the Defense Advanced Research Projects Agency (DARPA) under Agreement No. HR00112490483.
 Approved for public release; distribution is unlimited.
 TB and HO acknowledge support from the Vannevar Busch Fellowship program (ONR award number N000142512035).

\paragraph*{Author contributions:}
TB, AG, HO, and TS contributed to conceptualizing the approach, performing the research, and writing the article.


\paragraph*{Competing interests:}
There are no competing interests to declare.
\subsection*{Supplementary materials}
Materials and Methods\\
Supplementary Text\\
Figs. S1 to S2\\
Tables S1 to S4


\newpage


\renewcommand{\thefigure}{S\arabic{figure}}
\renewcommand{\thetable}{S\arabic{table}}
\renewcommand{\theequation}{S\arabic{equation}}
\renewcommand{\thepage}{S\arabic{page}}
\setcounter{figure}{0}
\setcounter{table}{0}
\setcounter{equation}{0}
\setcounter{page}{1} 


\begin{center}
\section*{Supplementary Materials for\\ \scititle}

	Th\'{e}o Bourdais$^{1}$,
	Abeynaya Gnanasekaran$^{2}$,\\
    Houman Owhadi$^{1}$,
    Tuhin Sahai$^{2*}$\\
	\small$^{1}$California Institute of Technology, Pasadena, CA 91125, USA.\\
	\small$^{2}$SRI International, Menlo Park, CA 94025, USA.\\
	\small$^\ast$Corresponding author. Email: tuhin.sahai@sri.com\\
	\small Authors are listed alphabetically and contributed equally to this work.

\end{center}

\subsubsection*{This PDF file includes:}
Materials and Methods\\
Supplementary Text\\
Figs. S1 to S2\\
Tables S1 to S4


\newpage

\section{Materials and Methods}

\subsection{Ensemble MCTS and  RL for Algorithm Discovery}\label{SecMCTSsupp}
We adapt the AlphaZero variant of MCTS \cite{browne2012survey,silver2017mastering,silver2018general,schrittwieser2020mastering} as a ``babbling’’ engine that chains tokens to obtain candidate algorithms. We use our adapted RL approach to (i) find generic solvers for the QAP and construct quantum circuits for quantum search and optimization, (ii) construct instance-adapted algorithms that are generated on-the-fly.  Details of our ensemble MCTS approach follow, and we omit aspects that do not deviate from AlphaZero. 


\noindent\textbf{State Representation.}
At depth $d$, state $s_d$ is defined by the sequence of previous actions taken $(a_1,\dots,a_{d-1})$. Additional information can be extracted from the state to get a more informative representation to input the neural network, such as loss increments of the states $(\mathcal{L}(s_2)-\mathcal{L}(s_1),\dots,\mathcal{L}(s_d)-\mathcal{L}(s_{d-1}))$ and the sequence of time taken to compute each action (see below). 

\noindent\textbf{Transformer-Based policy and value networks.}
Having formulated the state as a sequence of actions, rather than a problem-specific state such as a permutation or a quantum state, it is natural to use transformers to perform the task of predicting the policy and value from the state. By abstracting the problem this way, we obtain a unified learning framework for algorithm discovery, independant of the underlying problem. It also ensures effective generalization across problem instances of varying sizes, as the tokens themselves are defined for any size.

\noindent\textbf{Ensemble Variant and Handling Stochasticity.}
To accommodate stochasticity arising from random initializations and stochastic actions, we use ensembles of candidate solutions at each node, typically 5 to 10, and keep track of the best candidate ever seen along each path. Any time a node is reached, we compute a new batch of candidate solutions using $s_d=(a_1,\dots,a_{d-1})$. We compute the Gibbs average of the losses ($\sum_i \mathcal{L}_ie^{-\beta\mathcal{L}_i}/\sum_i e^{-\beta\mathcal{L}_i}$) of all candidates ever seen at that node, which gives an estimate of the expected loss achieved by the sequence of action $s_d$, $\mathcal{L}(s_d)$. This gives more information of a node's quality than sampling one candidate, and is fed to the transformers. 

\noindent\textbf{Computational cost control.} 
To account for the vast difference in cost between different primitives, our MCTS search incorporates computational cost into its decision process. A total compute budget is allocated and halts the program once no further action is feasible within the budget. To account for the cost of the policy and value network evaluations, which are proportional to the depth of the tree, we also add a fixed compute penalty at each depth level. This compute budget information is also fed to the transformers, to allow greater planning and compute management. We also add an early stopping primitive, so that not all of the compute is necessarily expended. Combined with a discounting of the rewards for high use of compute, this fosters the discovery of computationally efficient algorithms.

\subsection{CLP applied to the QAP}

\subsubsection{Discovering new strategies with low-level primitives and CLP}\label{secs15}
Our algorithm discovery framework is capable of operating with both low-level and high-level primitives. Using the QAP as a prototypical example, we now describe how it discovers new algorithms by composing simple base primitives. The framework alternates between two phases: (1) learning to compose primitives using MCTS, and (2) introducing new primitives by concatenating existing ones, following a strategy inspired by Byte-Pair Encoding (BPE) \cite{gage1994new}.


 \paragraph{Computational grammar.}
When analyzing existing algorithms, we observe that they often share common building blocks, which we will interpret as low-level primitives. Many of these primitives naturally correspond to matrix-level operations, such as gradient computations. However, these intermediary computations often impose specific input requirements, such as being a permutation, that are not guaranteed to be preserved in their outputs. This imposes constraints on how primitives can be chained.
Furthermore, certain fundamental algorithmic constructs do not neatly map to operations at the matrix level. For instance, a for-loop does not itself directly operate on matrices; instead, it represents repeated application of another primitive. To address this, we introduce the notion of special primitives, operators that act upon other primitives to generate new composite primitives. Because the effect of a special primitive depends on the structure of the primitive it modifies, additional constraints arise when composing them. All these constraints naturally give rise to a grammar that defines the set of valid primitive compositions.
There is an inherent trade-off between the expressivity afforded by a richer primitive set and grammar, and the complexity involved in effectively composing these primitives. To manage this trade-off, we restrict special primitives to operate exclusively on a single primitive at a time.
As a result, certain algorithms, such as Frank-Wolfe, while expressible in principle, are not immediately accessible due to their reliance on nested applications of special primitives. In such cases, expanding the primitive set through Byte-Pair Encoding becomes essential to recover complex algorithms. 

\paragraph{Used tokens.}
The chosen primitives are identity ($x\mapsto x$), gradient (GRAD: $x\mapsto\nabla\mathcal{L}(x)$), Linear Sum Assignment (LSA: $x\mapsto\arg\min_{\pi}\langle x,\pi\rangle_F$), and negative (NE: $x\mapsto -x$). Special primitives include a $k$-iteration loop (FOR) and residual updates (RU: $x+P(x)$). We also introduce two primitives that generate multiple permutations, e.g. sampling multiple random permutation, or considering all permutation that are one 2-city flip away from the current permutation, apply a given permutation, and keep the resulting state that has the smallest loss. Mathematically, they are (PU: $\arg\min\mathcal{L}(P(y_i))$, $y_i\sim\mathcal{U}$), and parallel 2-city flips (2SWAP: $x\mapsto\arg\min\{\mathcal{L}[P(y_{i,j})]\ |\ y_{i,j}=SWAP(x,i,j)\}$, where $SWAP(x,i,j)$ is the $x$, with indices $i$ and $j$ swapped). These two primitives allow for effective complexification, for instance the 2-city flip can produce $k$-optimal moves, including 2-OPT and $k$-OPT by chaining. Also, recall that a STOP primitive is available, to learn to perform early stopping (see computational cost control in \ref{SecMCTSsupp}).


 \paragraph{Algorithmic Byte-Pair Encoding (A-BPE).}

 After each round of babbling, we collect the sequences of tokens selected by ensemble MCTS. We count how often each pair of consecutive tokens appears in the corpus, excluding pairs that reconstruct existing tokens (e.g., applying the negative primitive twice to form the identity). The most common pair is merged into a new token, and all its instances in the corpus are replaced accordingly. This process repeats until no remaining pair occurs at least 10 times.

 \paragraph{Results.}
We run multiple rounds of consecutive babbling, with 10 iterations of self-play each, and A-BPE on QAP problems of size 15 then 40. The results of this experiment are shown in figure \ref{fig:low_level_primitives}. After one round of babbling, the majority of rollouts end with a 2-city swap and identity, i.e. the best flip (denoted 2FLIP, defined as $2SWAP\circ ID$). After one round of A-BPE concatenates 2FLIP into a single token, the composition of 2FLIP with a for loop becomes feasible, thereby, yielding the popular 2OPT strategy that is used in multiple algorithms such as the Lin-Kernighan heuristic. While around 60\% rollouts finish with 2OPT-\textgreater{}GRAD-\textgreater{}LSA, the success rate (i.e. the frequency of finding the global optimum) reaches 80\%, indicating that other, instance-specific strategies are discovered. After the second round of A-BPE, we move to problems of size 40, which is significantly harder, and renders 2OPT impractical due to its much higher computational cost and reduced effectiveness. Two rounds of A-BPE expands the vocabulary to 107 tokens, which include parts of existing and novel algorithms such as,  2SWAP, 3FLIP (i.e. 2SWAP(2SWAP(ID)), FOR(GRAD-\textgreater{}LSA), GRAD-\textgreater{}LSA-\textgreater{}FLIP, LSA-\textgreater{}GRAD, NE-\textgreater{}LSA.

In particular, A-BPE enables the discovery of the previously unavailable Frank-Wolfe algorithm (NE-\textgreater{}LSA-\textgreater{}GRAD-\textgreater{}FOR[RE(LSA-\textgreater{}GRAD)]-\textgreater{}LSA). Note that NE-\textgreater{}LSA is an $L_2$ projection to the space of permutations, allowing the algorithm to be applied to any initial matrix. While this strategy becomes quickly dominant, the success rate takes more time to grow. We observe the RL algorithm learns to perform random restarts, using a large number of PU(GRAD-\textgreater{}LSA) before FW. At the final iteration of self-play, we reach 93\% success rate, as new strategies and instance-specific adaptations are learned. Overall, these dynamics demonstrate not only the automatic rediscovery of classical heuristics but also the progressive invention of ever more effective, problem-tailored algorithms.

\subsubsection{Improving algorithm hyperparameters using CLP}\label{sec:cycliclr}
\begin{figure}[h]
\centering
\includegraphics[width=\linewidth]{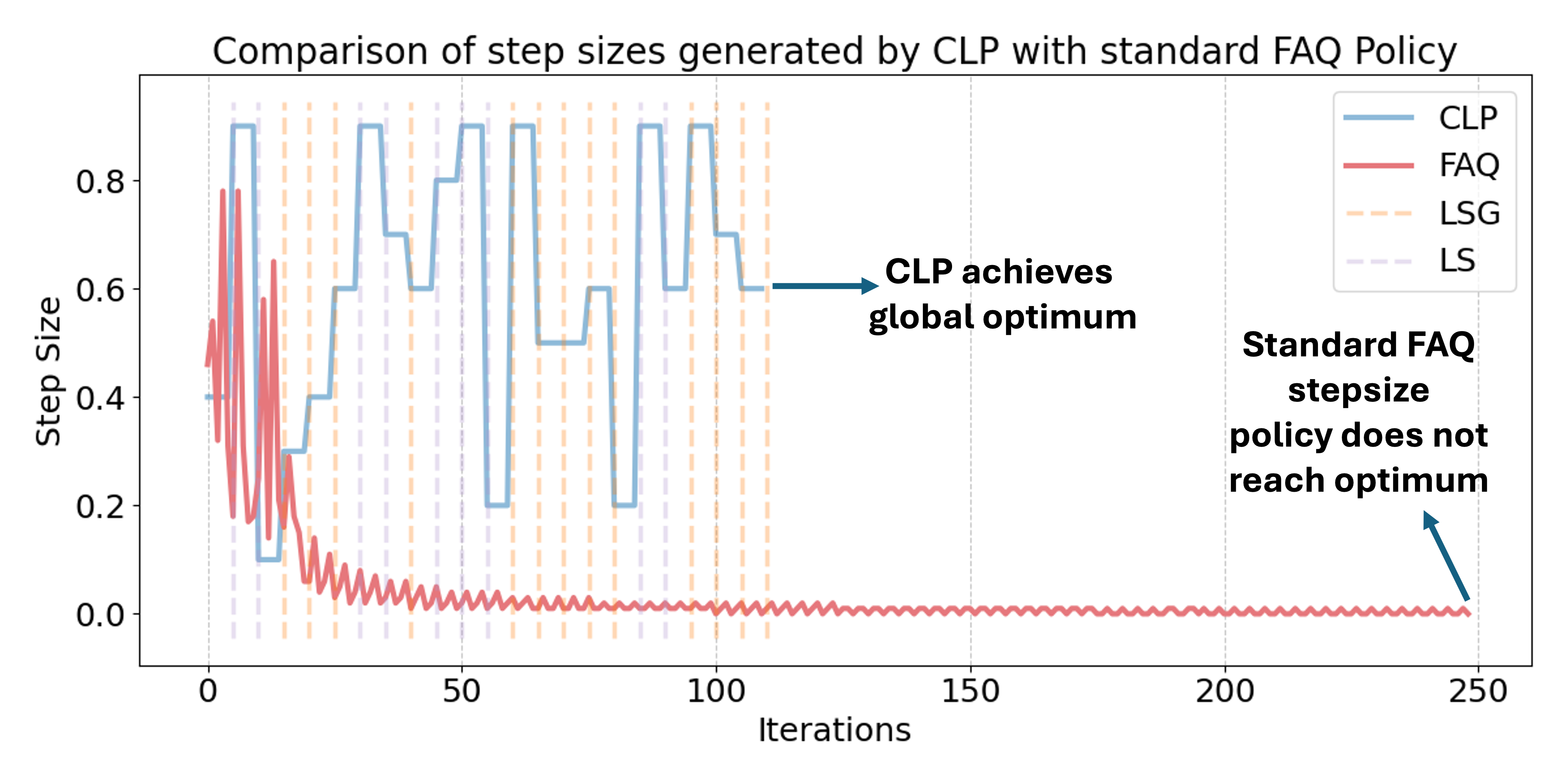}
\caption{Comparison of the optimal step sizes for the FAQ algorithm generated by our CLP approach compared to the standard step size schedule implemented in \texttt{scipy}. CLP discovers a cyclic rate (blue) that reaches global optimum unlike the standard line search-based approach used in \texttt{scipy} (red)}
\label{figcycling}
\end{figure}

In this experiment, we optimize the learning rate of the Frank-Wolfe algorithm. 
Recall that the FW algorithm for QAP iteratively updates:
\[
P_{k+1} = (1-\gamma_k)P_k + \gamma_k\LSA(\nabla\mathcal{L}(P_k)),
\]
where $\gamma_k$ is a tunable step-size that typically follows $\frac{2}{2+k}$. Using a line search~\cite{vogelstein2015fast} at each step yields a similar performance and decay in the learning rate. In order to optimize this schedule, we discretize the interval $[0,1]$ with 20 equidistant values. We define our computational tokens to be the 20 different choices of learning rate. Using CLP, we optimize the chaining of these tokens, creating learning rate schedules that are tailored to each instance. \\
Automatic selection via CLP reveals that cyclical step-size schedules (as in Fig.~\ref{figcycling}) yield improved performance and an ability to reach global optimum in tested instances. Notably, this phenomenon mirrors cyclical learning-rate strategies in neural networks, which achieve better accuracy and faster convergence without extensive tuning~\cite{smith2017cyclical}.

\subsubsection{Reaching state-of-the performance using high level primitives and CLP}\label{sec:high_level_primitives}

We define the high-level primitives as existing heuristics used to solve the QAP. They are simulated annealing, Frank-Wolfe (also known as Fast Approximate QAP) with 2 variants to project from doubly stochastic matrices to permutations, 2OPT, 3OPT and two variants of orthogonal optimization, with two variants as well as the early stopping primitive (see computational cost control in \ref{SecMCTSsupp}). Note that the early stopping is not used in small instances (size less than 25).\\
One pair of models, policy and value networks, is trained on examples generated from CQAP and PQAP, of varying size from 10 to 80. We compare our results to simply using Simulated Annealing for many more iterations. This is a simple baseline, which is known to converge to the true solution if given an exponentially large number of steps \cite{peng1996simulated}. We also code a simple branch-and-bound strategy, using the Gilmore-Lawler bound, with a limit on the total time spent. Finally, for a stronger benchmark, we use the commercial Gurobi solver by interpreting the QAP as a mixed-integer quadratic problem, and setting the runtime to be the same as our method. All examples up to size 80 run in under 200s, with the longest run taking less than 30 minutes for a size of 150.

\paragraph{Composite QAP (CQAP) Generator.} The \emph{composite} QAP (CQAP) generator~\cite{drugan2015generating} produces problems with varying structural complexity, facilitating robust evaluation of algorithmic performance. Our framework achieves optimality in all tested instances. In comparison, other methods get non optimal results, with a gap that widens as the size of the problem increases. Observing no gap between training and testing performance, we can conclude our framework has learned effective strategies for the whole CQAP class of problems. 

\paragraph{Palubeckis QAP (PQAP) Generator.}
The Palubeckis QAP (PQAP) generator~\cite{palubeckis2000algorithm} represents a sophisticated method for crafting challenging QAP instances with precisely known optimal solutions. 
Through Hamming distance analysis, we observed that the PQAP optimization landscape is notably more challenging compared to CQAP. It features steep and narrow valleys around global optima and a high density of local optima, complicating heuristic search efforts significantly.\\
Despite this difficulty, our achieves 1\% optimality gap in all tested instances, and beats the other methods in all but one example. Once again this performance is consistent with training performance, highlighting the capability of our method to generalize to other instances in the training distribution.

\paragraph{Benchmarking on QAPLIB.}
We evaluate our method using the widely recognized QAPLIB~\cite{qaplib} library, known for its challenging benchmark problems and completely unseen in the training of our method. CLP consistently achieves state-of-the-art performance, achieving the best-known objective values across multiple tested instances. It beats or equals other methods in most instances. Crucially, Gurobi beats our method in only one test problem of size larger than 25. We also note that global optimum is achieved for problems of size up to 100. This shows the excellent generalization of our method to unseen problems, with potentially different structure from the CQAP and PQAP generators used in training. Detailed comparative results are provided in Table~\ref{tab:qaplib_comparison}.

\subsubsection{Using CLP to manually create algorithms}\label{secs14}

While we present a method for automated algorithm discovery, CLP can be a useful framework for manually designing algorithm. In the process of developing our framework, we went through the steps of tokenizing, complexifying and improving by trial and error an algorithm, giving us valuable insight on how to automate this process. It resulted in an algorithm capable of achieving state-of-the-art performance with minimal computational resources when compared to other heuristics. While this algorithm was ultimately not implemented in automated discovery, it gives valuable insight on how high-level complexification within CLP is abstracted away from direct code-generation. Consequently, it fundamentally differs
from
approaches~\cite{lehman2023evolution, romera2024mathematical, liu2024evolution, novikov2025alphaevolve}. 

Starting from the Frank-Wolfe algorithm, which is a gradient descent in the space of doubly stochastic matrices, we observed that projecting the final doubly stochastic matrix to a permutation could prove to be difficult. While using $L_2$ projection or the permutation of steepest descent gives good results, taking the best of the two improves results most. This yielded the best of both token.  The gradient descent would often get stuck in local minima, as the optimization is not necessarily convex. To address this, we introduce a biasing token that adds a repulsive penalty to the loss. Specifically, given the current doubly stochastic matrix $x_0$ and a biasing matrix $T$, the biased loss is $\mathcal{L}_B(x)=\mathcal{L}(x)+\gamma \operatorname{Tr}(x^T  T)$. This allows one to push the gradient descent away from the current solution. To introduce some variety in the exploration, we introduced another special primitive similar to the parallel states primitive in \ref{secs15}. From a permutation $x$, take 10 uniform random permutations, $x_i$, and 10 random scalars $U_i\sim\mathcal{U}([0,1])$. Letting $y_i=P(U_ix_i+(1-U_i)x)$ for $P$ an arbitrary token acting on doubly stochastic matrices, this parallel state with perturbation token picks $\arg\min \mathcal{L}(y_i)$ as its output. This is similar to implementing a deflation technique to discover new solutions~\cite{farrell2016computation}. This variety allows us to compute a Gibbs average, $y^{\beta}=\frac{\sum_i y_i e^{-\beta\mathcal{L}(y_i)}}{\sum_i e^{-\beta\mathcal{L}(y_i)}}$. It extracts information about the most likely values for the true optimal on average, and is used to define the bias token, $T=y^{\beta}\odot(1-y^{\beta})\odot x_0$ (writing $\odot$ for the entrywise product).\\
Having complexified our vocabulary with these new tokens, we manually crafted a highly performant algorithm that solved all tested instances of the CQAP generator up to size 160. This early success motivated us to extend and automate this approach, using the CLP framework. 

\subsection{Discovering a new quantum circuit for Grover's algorithm with lower depth using CLP}\label{SecsuppGrover}

The \texttt{GroverGame} initializes with $n$ qubits and four primitive operations. An oracle randomly selects a target state from the $2^n$ possible states, with all states permuted each run to ensure generalization across arbitrary targets. The iteration limit is set to Grover's theoretical bound, $\lceil\frac{\pi}{4}\sqrt{N}\rceil$, preventing the discovery of algorithms that exceed it. The game terminates either in \emph{success} ($+1$), when a gate sequence identifies the target state within a tolerance of $3\times10^{-2}$, or in \emph{failure} ($-1$), when the maximum allowed sequence length is reached without success.


    

Upon termination, training data for the neural network is generated at each step, consisting of an observation tensor (quantum states), the overlap between current and target states, chosen primitives, and a modified reward value. To improve learning efficiency and encourage minimal circuit depth, raw rewards ($\pm 1$) are adjusted: for successful cases, shorter gate sequences are rewarded using $v = 1 - \frac{\text{len(sequence)}}{\text{max\_iterations}}$, whereas failures receive partial credit based on correctly identified targets (states), calculated as $v = -1 + \frac{\#\,\text{targets achieved}}{N}$. This modification incentivizes efficiency and mitigates sparse-reward problems common in reinforcement learning. Early-stage training faces sparse positive examples, leading to batch balancing of positive and negative samples (approximately 50\% each) via sampling with replacement when necessary. This ensures effective learning of rare successful gate sequences.

\paragraph{Analysis.}
We now demonstrate that our optimized gate sequence shown in Fig.~\ref{fig:grover_comparison} is mathematically equivalent to Grover's algorithm  but more efficient in gate usage. 

First, we summarize the \emph{standard Grover algorithm}. Start from the state $|0\rangle^{\otimes n}$ and apply Hadamard gates to achieve the uniform superposition state
\[
|s\rangle = H^{\otimes n}|0\rangle^{\otimes n} = \frac{1}{\sqrt{N}}\sum_{x=0}^{N-1}|x\rangle.
\]
Then, for $O(\sqrt{N})$ iterations, repeat two operations: (i) apply the oracle $U_f$ that marks the target state $|w\rangle$ by flipping its phase, and (ii) apply the diffusion operator $D = 2|s\rangle\langle s| - I$.
Analyzing one iteration, after applying the oracle $U_f$, we have:
\[
U_f|s\rangle = |s\rangle - \frac{2}{\sqrt{N}}|w\rangle.
\]
Next, applying the diffusion operator gives:
\[
DU_f|s\rangle 
= 2|s\rangle\langle s|\left(|s\rangle - \frac{2}{\sqrt{N}}|w\rangle\right) - \left(|s\rangle - \frac{2}{\sqrt{N}}|w\rangle\right).
\]
Noting that $\langle s|w\rangle = 1/\sqrt{N}$, this simplifies  to:
\[
DU_f|s\rangle = \left(1 - \frac{4}{N}\right)|s\rangle + \frac{2}{\sqrt{N}}|w\rangle.
\]

The \emph{optimized Grover algorithm} proceeds as follows. Starting from $|0\rangle^{\otimes n}$, apply $X$ gates to all qubits, resulting in the state $|1\rangle^{\otimes n}$. Next, apply Hadamard gates to form a modified superposition:
\[
|s'\rangle = H^{\otimes n}|1\rangle^{\otimes n} = \frac{1}{\sqrt{N}}\sum_{x=0}^{N-1}(-1)^{|x|}|x\rangle,
\]
where $|x|$ is the Hamming weight (number of ones) in the binary representation of $x$. Then, iteratively apply the oracle $U_f$ to mark the target state and use the modified diffusion operator:
\[
D' = 2|s'\rangle\langle s'| - I,
\]
repeating these steps for $O(\sqrt{N})$ iterations.

       
       

We now analyze the optimized circuit's effect on the initial state $|0\rangle^{\otimes n}$. Applying $X$ gates transforms it to $|1\rangle^{\otimes n}$, followed by Hadamard gates yielding the modified superposition:
\[
|s'\rangle = \frac{1}{\sqrt{N}}\sum_{x=0}^{N-1}(-1)^{|x|}|x\rangle.
\]
The oracle $U_f$ acts similarly as before, marking the target state $|w\rangle$:
\[
U_f|s'\rangle = |s'\rangle - \frac{2}{\sqrt{N}}(-1)^{|w|}|w\rangle.
\]
The modified diffusion operator is defined as:
\[
D' = 2|s'\rangle\langle s'| - I = H^{\otimes n}(2|1\rangle\langle 1| - I)H^{\otimes n},
\]
which can be efficiently implemented using the gate sequence $H^{\otimes n}\rightarrow\text{MCZ}\rightarrow H^{\otimes n}$, as illustrated in Fig.~\ref{fig:grover_comparison}.
Applying $D'$ after $U_f$ gives:
\[
D'U_f|s'\rangle 
= \left(1 - \frac{4}{N}\right)|s'\rangle + \frac{2}{\sqrt{N}}(-1)^{|w|}|w\rangle.
\]
Notably, this result differs from standard Grover’s algorithm by only a global phase factor $(-1)^{|w|}$, which does not affect measurement probabilities. Therefore, our optimized circuit remains mathematically equivalent. The crucial advantage is that initializing with $|1\rangle^{\otimes n}$ directly simplifies the diffusion operator, removing the intermediate $X$ gates required by the standard implementation.

 \paragraph{Comparison to Related Work.}
Gilliam et al.~\cite{gilliam2020optimizing} reduce circuit depth in Grover's algorithm by replacing Hadamard gates with \( RX(\pi/2) \) rotations, eliminating extra \( X \) gates (see Table \ref{tab:grover-tech-comparison}). However, we note that $RX$ gates are challenging to implement on real-world hardware platforms. Our circuits, on the other hand, are more practical. For example, our circuits are easily implementable on photonic based quantum computing solutions by removing the need for $RX$ gates. Our approach achieves the same depth reduction using only \( H \) and \( X \) gates, preserving compatibility with Clifford-based frameworks and providing a more conventional construction. Wu et al.~\cite{wu2023circuit} and Zhang and Korepin~\cite{zhang2020depth} propose multistage variants that reduce depth via hierarchical partitioning and local diffusion operators. Piron et al.~\cite{piron2025mixed} introduce a hybrid strategy that reduces Oracle calls by combining fewer Grover iterations with bounded classical retries. In contrast, our method avoids block partitioning and classical feedback, and achieves depth reduction via a simple basis transformation. We also note that in future work, we plan to include hardware implementation details to design circuits that are tailored to the quantum platform of interest.

\renewcommand{\arraystretch}{0.4}
\begin{table}[h]
\centering
\small
\caption{Comparison of Grover Optimization Strategies}
\label{tab:grover-tech-comparison}
\begin{tabular}{lcccc}
\toprule
\textbf{Method} & \textbf{Diffusion} & \textbf{Approximation?} & \textbf{Hybrid?} & \textbf{Depth} \\
 & \textbf{Operator} &  &  & \textbf{Reduction} \\
\midrule
\textbf{Ours} & $H \cdot \text{MCZ} \cdot H$ & No & No & $2\times$ fewer 1-qubit gates \\
Gilliam et al.~\cite{gilliam2020optimizing} & $RX \cdot \text{MCZ} \cdot RX$ & No & No & $2\times$ fewer gates per iteration \\
Wu et al.~\cite{wu2023circuit} & Hierarchical  & No & Yes (2-stage) & $1.2\times$ depth reduction \\
Zhang \& Korepin~\cite{zhang2020depth} & Local diffusion & Yes & No & $20\%$ empirical reduction\\
Piron et al.~\cite{piron2025mixed} & Standard & No & Yes & $10\%$ reduction \\
\bottomrule
\end{tabular}
\end{table}

\subsection{CLP-based QAOA results}
\label{sec:clpqaoa}

\paragraph{The Quantum Approximate Optimization Algorithm (QAOA)} \cite{farhi2014quantum} is a hybrid quantum-classical optimization algorithm designed to approximately solve combinatorial optimization problems formulated as a bit-string search $z \in \{0,1\}^n$ that minimizes a given cost function $C(z)$. The algorithm was designed for the current generation of noisy quantum platforms with the aim of demonstrating their benefit and has been applied to problems in graph theory, supply chain optimization, task scheduling, and energy management to name a few.
By encoding the cost function in a  Hamiltonian form:
$
H_C = \sum_{z \in \{0,1\}^n} C(z)\,|z\rangle \langle z|$, 
the algorithm approximates solutions by preparing and optimizing parameterized quantum states $|\psi(\gamma,\beta)\rangle$, which depend on adjustable parameters $(\gamma, \beta)$. These parameters are tuned using classical optimization routines such as gradient descent. These quantum states are implemented on quantum devices using quantum circuits that consist of alternating layers: a \emph{cost Hamiltonian layer} ($e^{-i\gamma H_C}$), that encodes the cost function, and a \emph{mixer Hamiltonian layer} ($e^{-i\beta H_M}$), that are designed to facilitate exploration of the energy landscape. Therefore, one can map the initial state $|+\rangle^{\otimes n}$ to the final quantum state using:
$|\psi(\gamma, \beta)\rangle = e^{-i\beta_p H_M} e^{-i\gamma_p H_C} \dots e^{-i\beta_1 H_M} e^{-i\gamma_1 H_C}|+\rangle^{\otimes n}$. As mentioned previously, the optimal parameters $(\gamma_i, \beta_i)$ are computed by classically minimizing the expectation  of the cost Hamiltonian:
\[
(\gamma^*, \beta^*) = \arg\min_{\gamma,\beta} \langle \psi(\gamma,\beta)| H_C |\psi(\gamma,\beta)\rangle.
\]

\begin{figure}[h]
    \centering
    \includegraphics[width=1\linewidth]{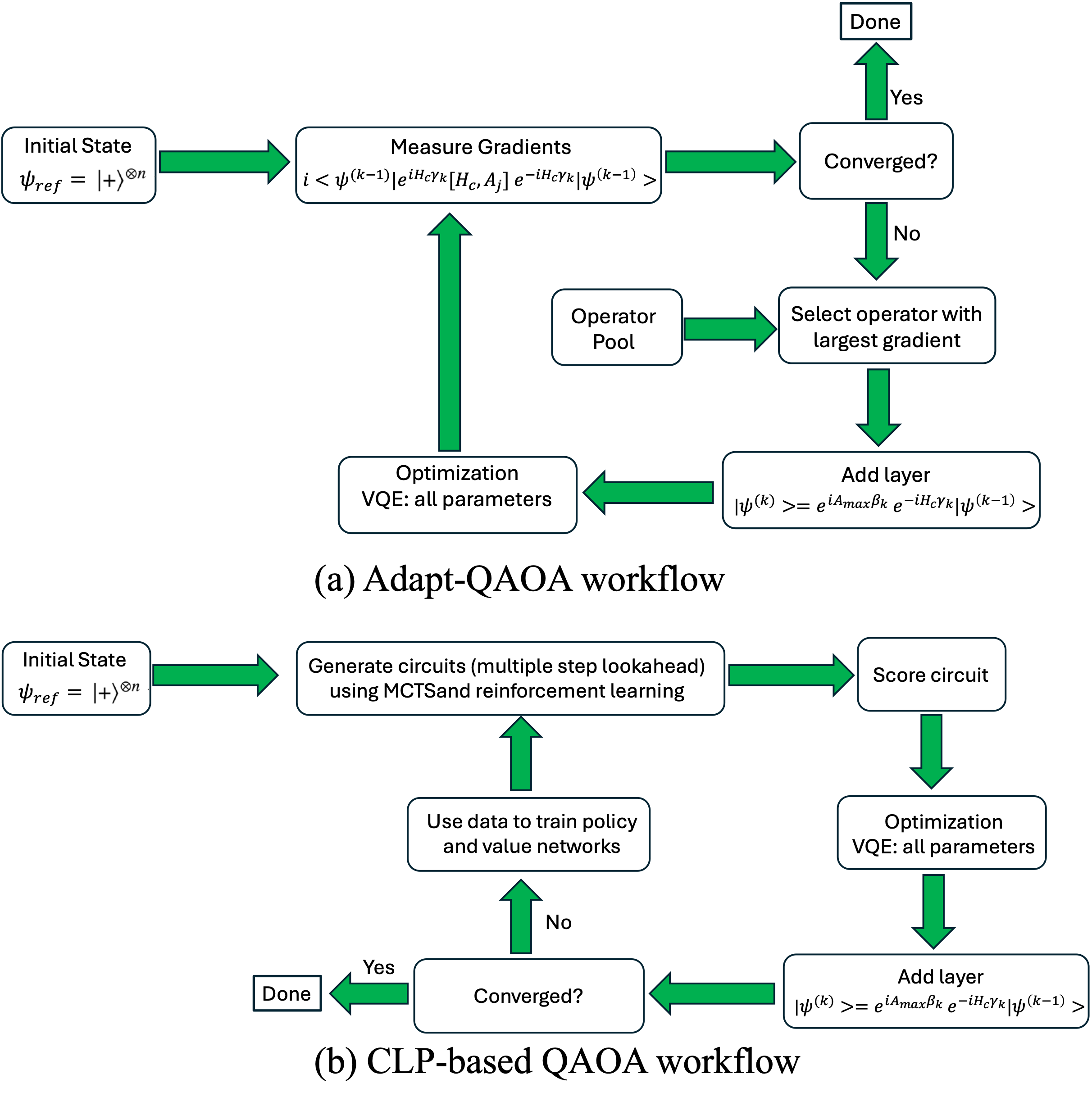}
    \caption{Comparison of workflows used by ADAPT-QAOA and our CLP-based approach for generating quantum circuits for QAOA.
}
    \label{fig:workflowQAOA}
\end{figure}

\paragraph{Results on the Quantum Approximate Optimization Algorithm (QAOA)}
An important challenge within the QAOA setting is the design of shallow mixer Hamiltonian quantum circuits to minimize the optimization cost. We focus our efforts on using our CLP framework for designing the mixer Hamiltonian circuits for QAOA. We compare our approach with an adaptive version of QAOA known as ADAPT-QAOA~\cite{zhu2022adaptive}. The approach is a standardized solution implemented within Nvidia's quantum computing library (CUDA-Q). ADAPT-QAOA is an iterative classical-quantum algorithm that adaptively builds problem-specific mixer Hamiltonians for solving a given combinatorial optimization problem. As shown in Fig.~\ref{fig:workflowQAOA} (a), unlike standard implementations of QAOA that use the \emph{same} mixer Hamiltonian at each step, ADAPT-QAOA iteratively selects operators from a pool of candidate gates based on their gradient contributions to the cost function. The mixer that produces the \emph{steepest descent} at each step is selected and appended to the circuit  ansatz. All parameters ($\gamma$, $\beta$) of the cost and mixer Hamiltonians are then optimized classically using variational quantum eigensolver (VQE). The process continues until either a convergence criteria is met or the user prescribed maximum circuit depth is reached.

Our CLP-based approach significantly improves the ADAPT-QAOA approach by learning to propose the optimal problem-specific mixer Hamiltonian circuit at each step. Unlike ADAPT-QAOA that uses steepest descent to iteratively add successive gates to the circuit ansatz, our CLP-based QAOA framework, as depicted in Fig.~\ref{fig:workflowQAOA} (b), simultaneously considers entire chains of gates for the mixer circuits by using using MCTS and reinforcement learning to implement a \emph{lookahead} policy. 

To test our approach, we use the CLP framework to efficiently construct the mixer Hamiltonian circuits for instances of the MAX-CUT problems on $p$-regular graphs that \emph{reduce the error for a user prescribed maximum circuit depth}. Our primitives consist of the same gates used with the ADAPT-QAOA framework for generating circuits~\cite{zhu2022adaptive} . Namely, we use the standard QAOA mixer pool~\cite{zhu2022adaptive} that consists of $\sum_{i\in Q}\{X_i\}$, single-qubit $\cup_{i\in Q}\{X_i\}$, and multi-qubit entangling gates $\cup_{i, j \in Q\times Q} \{B_iC_j|B_i, C_j \in \{X, Y, Z\}\}$, where $Q$ is the set of qubits. We note that only Pauli strings that have an even number of $Y$ or $Z$ gates are retained in our mixer pool~\cite{zhu2022adaptive}. We generate 1000 $p$-regular graphs for the training of our neural networks and limit the depth of the mixer circuits. We then test the CLP framework on 21 graphs that were not seen in training. The CLP-framework outperforms ADAPT-QAOA in 100\% of test cases and corresponds to an average improvement of $34.62\%$. The results are tabulated in table~\ref{tab:clpqaoa}.



\begin{table}[ht]
\centering
\caption{Performance Comparison: CLP-QAOA vs. ADAPT-QAOA and \% improvement}
\label{tab:clpqaoa}
\renewcommand{\arraystretch}{1.0}
\setlength{\tabcolsep}{2pt}
\arrayrulecolor{lightgray}\begin{tabular}{|c|c|c|c|c|c|c|c|c|c|c|c|c|c|c|c|c|c|c|c|c|c|}
\hline
\textbf{Circuit} & 1 & 2 & 3 & 4 & 5 & 6 & 7 & 8 & 9 & 10 & 11 & 12 & 13 & 14 & 15 & 16 & 17 & 18 & 19 & 20 & 21 \\
\hline
\textbf{CLP} & \small-2.8 & \small-2.6 & \small-3.2 & \small-4.1 & \small-3.7 & \small-2.3 & \small-2.8 & \small-2.4 & \small-3.9 & \small-2.9 & \small-2.4 & \small-2.8 & \small-3.3 & \small-2.9 & \small-3.5 & \small-3.3 & \small-3.4 & \small-2.6 & \small-3.3 & \small-4.1 & \small-3.3 \\
\hline
\textbf{ADAPT} & \small-2.1 & \small-2.1 & \small-2.2 & \small-3.2 & \small-2.5 & \small-1.5 & \small-2.0 & \small-2.1 & \small-2.6 & \small-2.0 & \small-2.1 & \small-2.3 & \small-2.4 & \small-2.2 & \small-2.6 & \small-2.6 & \small-2.4 & \small-2.1 & \small-2.7 & \small-2.5 & \small-2.4 \\
\hline
\textbf{\% Improve} & \small36 & \small23 & \small44 & \small30 & \small45 & \small49 & \small41 & \small16 & \small51 & \small44 & \small17 & \small19 & \small34 & \small35 & \small35 & \small28 & \small38 & \small21 & \small23 & \small61 & \small38 \\
\hline
\end{tabular}
\arrayrulecolor{black}
\end{table}

\section{Supplementary Text}

\subsection{List of major algorithms discovered over time}
The following table summarizes significant algorithms discovered throughout history, spanning mathematics, optimization, cryptography, and machine learning.

{\small
\begin{center}
\begin{longtable}{|l|c||l|c|}
\hline
\textbf{Algorithm} & \textbf{Year} & \textbf{Algorithm} & \textbf{Year} \\
\hline\hline
Ancient Egyptian Multiplication & 3000 BCE & Babylonian Square Root Approximation & 2000 BCE \\
Mesopotamian Algorithm for Square Roots & 1800 BCE & Euclidean Algorithm & 300 BCE \\
Method of Exhaustion (Archimedes) & 250 BCE & Sieve of Eratosthenes & 200 BCE \\
Chinese Remainder Theorem & 250 CE & Gaussian Elimination & 263 CE \\
Chakravala Method & 628 CE & Al-Khwarizmi’s Algebra & 820 CE \\
Al-Kindi's Cryptography & 850 CE & Cubic/Quartic Solving & 1540 \\
Logarithms & 1614 & Newton-Raphson & 1671 \\
Calculus & 1687 & Euler Numerical Methods & 1700 \\
Linear Regression & 1700 & Gradient Descent & 1800 \\
Ada Lovelace’s Algorithm & 1842 & Krylov Algorithm & 1900 \\
Cholesky Decomposition & 1900 & Turing Machine & 1936 \\
Finite Element & 1943 & Merge Sort & 1945 \\
Breadth First Search & 1945 & Simplex Algorithm & 1947 \\
Markov Chain Monte Carlo & 1948 & Matrix Decompositions & 1951 \\
Huffman Coding & 1952 & Metropolis Algorithm & 1953 \\
Hashing Algorithm & 1953 & k-means Clustering & 1956 \\
Fortran Compiler & 1956 & Perceptron & 1959 \\
QR Decomposition & 1959 & Dijkstra’s Shortest Path & 1960 \\
Branch and Bound & 1960 & Kalman Filter & 1960 \\
Quicksort & 1961 & Viterbi's algorithm & 1968 \\
Fast Fourier Transform & 1968 & A* Pathfinding & 1969 \\
Genetic Algorithms & 1970 & Bloom Filter & 1970 \\
Lin Kernighan Heuristic & 1973 & Strassen’s Matrix Multiplication & 1976 \\
Buchberger's algorithm & 1976 & Diffie–Hellman Key Exchange & 1977 \\
RSA Cryptosystem & 1977 & Expectation Maximization & 1977 \\
Lempel-Ziv Compression & 1977 & Beam Search & 1977 \\
Integer Relation Detection & 1977 & KMP algorithm & 1977 \\
Simulated Annealing & 1980 & Lattice basis reduction algorithm & 1982 \\
Ellipsoid Algorithm & 1984 & Reinforcement Learning & 1984 \\
Interior Point Algorithm & 1984 & Elliptic Curve Cryptography & 1984 \\
Fast Multipole & 1985 & Backpropagation for Neural Networks & 1986 \\
Convolutional Neural Networks & 1987 & JPEG Encoding & 1992 \\
Turbo Code & 1993 & Shor's Quantum Factoring & 1994 \\
Support Vector Machine & 1995 & Grover's algorithm & 1996 \\
PageRank & 1998 & BitTorrent & 2001 \\
Distributed Hash Table & 2001 & AKS Primality Test & 2002 \\
Packrat parser & 2002 & Community Detection Algorithms & 2003 \\
Khyber & 2003 & Sparse Signal Processing & 2004 \\
Monte Carlo tree search (MCTS) & 2006 & Bitcoin \& Blockchain & 2009 \\
Fully Homomorphic Encryption & 2009 & Lattice based cryptography & 2013 \\
Secure NTRU & 2013 & Deep Q Network & 2013 \\
Word2vec & 2013 & You Only Look Once & 2013 \\
Raft algorithm & 2013 & Transformer/Attention Algorithms & 2017 \\
Zk-STARK & 2021 & & \\
\hline
\end{longtable}
\end{center}
}

\subsection{Why is the QAP hard?}\label{Secqaphard}
The QAP \cite{koopmans1957assignment, loiola2007survey} is a strongly NP-hard problem \cite{garey1979computers}, meaning that not only does no known polynomial-time algorithm exist to solve the QAP (and it is unlikely that such an algorithm will ever be found unless $P=NP$), but also that constructing a constant-factor approximation algorithm is itself NP-hard. The QAP is often regarded as the ``hardest of the NP-hard problems'' due to its extreme computational difficulty, even for relatively small instances (e.g., $n=20$) \cite{burkard1997qaplib}.
Furthermore, any problem in NP (nondeterministic polynomial time) can be reduced to the QAP in polynomial time \cite{garey1979computers}. Thus, if we were able to solve the QAP efficiently (in polynomial time), we would be able to solve all NP problems efficiently, which would have profound implications for computational complexity theory.
The QAP also encompasses several well-known problems in combinatorial optimization.  For instance, by setting  $F_{i,j}=1$, $D_{j,i}=D_{i,j}\geq 0$, $D_{i,i}=0$ and $C_{i,j}=0$, the QAP simplifies to the Traveling Salesman Problem (TSP) \cite{applegate2006traveling}, which is a special case of the QAP and is generally considered much easier to solve. Moreover, by setting $C=0$ and allowing $D_{i,j}\in \{0,1\}$,
 the QAP reduces to the Graph Matching Problem \cite{livi2013graph}, a problem concerned with finding optimal node correspondences between two graphs. Finally, by setting both $F=0$ and $D=0$  the QAP simplifies to the LSA problem, which can be solved in polynomial time using the Hungarian Algorithm \cite{kuhn1955hungarian}.

\subsection{Why is the QAP important?}\label{Secqapimportant}

The optimization of logistics and supply chains, such as allocating resources to multiple sites during operations and determining inventory levels over time at various facilities, is inherently challenging due to the combinatorial scaling of search spaces and the dynamic evolution of states. These complexities often cause current algorithms to produce suboptimal solutions, with performance deteriorating significantly as problem size increases, potentially leading to undesirable outcomes ranging from increased operational costs to loss of life. 
In the context of supply chain management, QAP is particularly important for strategically placing facilities to minimize the time required to move items through a complex network. The problem's dynamic and uncertain nature, coupled with evolving structural changes over time, further complicates the optimization process. Existing methods frequently fail to adapt to these changes, underscoring the need for algorithms that can dynamically generate solutions in response to shifting conditions.
 In supply chain optimization, moderate-sized problems can involve over $\mathcal{O}(10^6)$ parameters, making the simulation and optimization of these networks inefficient with current algorithms. Additionally, other applications of QAP include facility layout and location (e.g., airport design), data center optimization, and very large-scale integration circuit design. Developing an automated algorithm discovery framework would therefore represent a significant industry and operational breakthrough, enabling the dynamic generation of algorithms that adapt to changing conditions and achieve lower-cost solutions by effectively relocating items to critical areas. Such advancements would not only enhance the efficiency and resilience of logistics but also offer substantial benefits across various sectors reliant on complex supply chain and facility optimization.

\subsection{Current methods for solving the QAP}
Current methodologies for solving the QAP span exact algorithms, heuristic/metaheuristic methods, and hybrid/emerging approaches,  detailed in  surveys \cite{cook1998combinatorial, loiola2007survey}.\\
\noindent\textbf{Exact methods} include systematic techniques such as Branch and Bound (B\&B), Cutting Plane methods, Dynamic Programming (DP), and Integer Programming (IP), providing guarantees of optimality but with significant computational demands for general instances.\\
\noindent\textbf{Heuristic algorithms} offer efficient approximations suitable for larger problems. Prominent examples include Simulated Annealing (SA), Genetic Algorithms (GA), Tabu Search (TS), Ant Colony Optimization (ACO), Particle Swarm Optimization (PSO), Iterated Local Search (ILS), and Greedy Randomized Adaptive Search Procedure (GRASP).\\
\noindent\textbf{Hybrid and emerging approaches} blend multiple methodologies to enhance performance. These include Memetic Algorithms (MA), Hybrid Genetic Algorithms (HGA), Parallel Algorithms, and novel techniques such as Quantum Algorithms and Machine Learning-assisted heuristics.

\subsection{Exhaustive list of primitives for the QAP}\label{sec:all_primitives}

\subsubsection{Low level primitives}

Low level primitives are tokens that do not directly solve the problem at hand, here the QAP, but can be chained to get an algorithm.\\
The following low level primitives are implemented in section \ref{secs15} are: \begin{itemize}
    \item \textbf{Identity} (ID): $x\mapsto x$
    \item \textbf{Linear Sum Assignment} (LSA): $x\mapsto\arg\min_{\pi}\langle x,\pi\rangle_F$
    \item \textbf{Negative} (NE): $x\mapsto -x$
    \item \textbf{Gradient} (GRAD): $x\mapsto\nabla\mathcal{L}(x)$
    \item \textbf{Early stopping} (STOP): No further computation, saves compute budget for higher reward. 
\end{itemize}
Additionally, \ref{secs15} uses special primitives, whose outputs are determined by the current state and next token $P$:\begin{itemize}
    \item \textbf{For loop} (usually 50 iterations) (FOR): $x,P\mapsto P\circ...\circ P(x)$
    \item \textbf{Residual update} (RU): $x,P\mapsto x+P(x)$
    \item \textbf{Parallel uniform} (PU): $x,P\mapsto \arg\min\mathcal{L}(P(y_i)),\ (y_i)_{i=1}^{10}\sim\mathcal{U}$
    \item \textbf{Parallel 2-city swaps} (2SWAP): $x,P\mapsto\arg\min\{\mathcal{L}[P(y^{i,j})]\ |\ y^{i,j}=SWAP(x,i,j)\}$, where $y^{i,j}_k = x_k$ if $k\neq i,j$, and $y^{i,j}_j=x_i$, $y^{i,j}_i=x_j$ is the permutation $x$, with indices $i$ and $j$ swapped
\end{itemize}
Finally, several primitives are considered for manual algorithm discovery \ref{secs14}: \begin{itemize}
    \item \textbf{Gibbs permutation average} (uses multiple permutations) (GIBBS): $(x_i)_{i=1}^k\mapsto \frac{\sum_i x_i e^{-\beta\mathcal{L}(x_i)}}{\sum_i e^{-\beta\mathcal{L}(x_i)}}$
    \item \textbf{Biasing tokens} (BIAS): Given a matrix $T$ and a doubly stochastic matrix $x_0$, changes the loss to $\mathcal{L}_b(x)=\mathcal{L}(x)+\gamma\operatorname{Tr}(x_0^T T)$
    \item \textbf{Best of both} (BEST): $x\mapsto \arg\min_{i\in\{1,2\}}(\mathcal{L}(y_i))$ where $y_1=LSA\circ NE(x)$ and $y_2 = LSA\circ GRAD (x)$
    \item \textbf{Parallel state with permutation} (PSP): $x,P\mapsto\arg\min\{\mathcal{L}[P(y_i)]\ |\ y_i=U_ix_i+(1-U_i)x\}$ where $x_i$ is a random uniform permutation, and $U_i\sim\mathcal{U}([0,1])$
\end{itemize}

\subsubsection{High level primitives}

The following algorithms are used to get the high level primitives results in \ref{sec:high_level_primitives}. Note that many of these primitives are rediscovered in \ref{secs15}.\begin{itemize}
    \item \textbf{Two-location optimal swaps} (2OPT): iteratively swaps two distinct index $i,j$ using the best possible swap, until no improvement is possible. Note that this is approximately achieved by $FOR(2SWAP\circ ID)$ (with enough iterations). 
    \item \textbf{Three-location optimal swaps} (3OPT): same as 2OPT, but a 3 location swaps can change up to 3 indices, $i,j,k$. Note that this can be approximately achieved using $FOR(2SWAP(2SWAP\circ ID))$. In practice, we limit the number of explorations of 3 city swaps, as this can be prohibitively expensive. 
    \item \textbf{Path reversal 2OPT} (P-2OPT): defining the path reversal such that $y^{i,j}=REVERSE(x,i,j)$, $y^{i,j}_k =x_{j+i-k}$ if $i\leq k\leq j$, else $y^{i,j}_k =x_{k}$, P-2OPT applies iteratively the best path reversal until no improvement. 
    \item \textbf{Path reversal 3OPT} (P-3OPT) choose the best 3 indices, $i,j,k$, and iteratively applies 2 path reversal, $REVERSE(REVERSE(x,i,j),j,k)$, until no improvement or maximum iteration number is reached.
    \item \textbf{Simulated annealing} (SA): This primitive performs $m$ steps of simulated annealing using two-location flips on an initial permutation $x$. At step $k$, simulated annealing picks two random indices, $i,j$, and performs $\tilde{x}^k=SWAP(x^{k},i,j)$. With probability $\min\left(1,\exp\left(-\frac{\Delta \mathcal{L}}{T_k}\right)\right)$ where $\Delta \mathcal{L}=\mathcal{L}(\tilde{x}^k)-\mathcal{L}(x_k)$, the move is accepted and $x^{k+1}=\tilde{x}^k$; else $x^{k+1}=x^k$. The temperature follows a decreasing schedule $T_k=T_0(1-\epsilon)^k$. This approach probabilistically accepts worse solutions, allowing escape from local minima, until $m$ iterations are completed.
    \item \textbf{Frank-Wolfe }(FW): Performs gradient descent in the space of doubly stochastic matrices. One iteration has the update $x^{k+1} = (1-\gamma_k)x^k+\gamma_k LSA(GRAD(x^k))$, where $\gamma_k=2/(2+k)$
    \item \textbf{Orthogonal relaxation with polar decomposition} (OP). Following the relaxation strategy described in~\cite{sahai2019continuous}, OP performs gradient descent in the space of orthogonal matrices using the Procrustes solution~\cite{sahai2019continuous, sahai2020dynamical}. Defining $\mathcal{Q}[x]$ as the Q factor of the QR-decomposition, the update rule is $x^{k+1}=\mathcal{Q}[x^{k}-\gamma^kGRAD(x^k)]$, with the learning rate $\gamma^k=0.5*0.95^k$. 
    \item \textbf{Orthogonal relaxation with polar decomposition and momentum} (OC): More complex update rule for gradient descent with momentum in the orthogonal matrices space, see \cite{sahai2019continuous, sahai2020dynamical}.
\end{itemize}
Note that the space of permutation matrices is exactly the intersection of the doubly stochastic matrices and orthogonal matrices, justifying the relaxations of both FW and OC.

\footnotesize
\renewcommand{\arraystretch}{0.4}
\setlength{\tabcolsep}{2pt}

\begin{longtable}{*{6}{c}!{\vrule width 1pt}*{6}{c}} 
  \caption{Relative gap ($(\mathcal{L}-\min\mathcal{L})/\min\mathcal{L}$, in \%) on QAPLIB Benchmark Instances (2 per row)} \label{tab:qaplib_comparison} \\
  \toprule
  \scriptsize Problem & \scriptsize Best Model & \scriptsize \textbf{CLP (ours)} (\%) & \scriptsize SA (\%) & \scriptsize BB (\%) & \scriptsize Gurobi (\%) &
  \scriptsize Problem & \scriptsize Best Model & \scriptsize \textbf{CLP (ours)} (\%) & \scriptsize SA (\%) & \scriptsize BB (\%) & \scriptsize Gurobi (\%) \\
  \midrule
  \endfirsthead

  \toprule
  \scriptsize Problem & \scriptsize Best Model & \scriptsize \textbf{CLP (ours)} (\%) & \scriptsize SA (\%) & \scriptsize BB (\%) & \scriptsize Gurobi (\%) &
  \scriptsize Problem & \scriptsize Best Model & \scriptsize \textbf{CLP (ours)} (\%) & \scriptsize SA (\%) & \scriptsize BB (\%) & \scriptsize Gurobi (\%) \\
  \midrule
  \endhead

  \midrule
  \multicolumn{12}{r}{{Continued on next page}} \\
  \midrule
  \endfoot

  \bottomrule
  \endlastfoot

  \begin{filecontents*}{QAPLIB_results.csv}
problem_name,best_model,rl_optimality_ratio_percent,sa_optimality_ratio_percent,bb_optimality_ratio_percent,gurobi_optimality_ratio_percent,problem_name_2,best_model_2,rl_optimality_ratio_percent_2,sa_optimality_ratio_percent_2,bb_optimality_ratio_percent_2,gurobi_optimality_ratio_percent_2
bur26f,CLP,0.0,1.0,8.6,11.5,nug18,CLP,0.0,4.7,21.9,1.2
bur26c,CLP,0.0,1.4,7.1,5.3,nug17,CLP,0.0,6.7,20.1,2.3
bur26e,CLP,0.1,1.4,8.1,10.9,nug16a,CLP,0.0,7.2,23.5,5.5
bur26h,CLP,0.1,2.1,8.8,9.8,nug15,CLP,0.0,6.1,19.5,3.1
bur26a,CLP,0.2,0.3,6.4,8.5,nug14,CLP,0.0,7.3,23.1,2.2
bur26b,CLP,0.2,0.7,6.7,7.3,nug12,CLP/Gurobi,0.0,5.2,15.6,0.0
bur26d,CLP,0.1,0.4,7.2,10.3,rou20,CLP,0.1,5.0,17.5,2.3
bur26g,CLP,0.3,0.7,8.5,10.2,rou15,CLP,0.0,5.3,21.3,6.2
chr25a,Gurobi,10.0,85.2,304.6,8.6,rou12,CLP/Gurobi,0.0,6.3,22.8,0.0
chr22b,CLP,1.6,30.4,192.6,7.1,scr20,CLP,0.0,12.4,59.0,2.2
chr22a,Gurobi,1.9,14.1,116.6,0.7,scr15,CLP/Gurobi,0.0,20.3,76.9,0.0
chr20b,CLP,7.0,56.5,213.7,11.3,scr12,CLP/Gurobi,0.0,6.0,45.1,0.0
chr20a,Gurobi,1.5,72.0,248.2,0.0,sko100e,CLP,0.0,9.0,16.4,17.3
chr20c,CLP/Gurobi,0.0,45.9,625.1,0.0,sko100c,CLP,0.1,8.4,17.6,16.7
chr18a,Gurobi,0.2,105.9,418.7,0.0,sko100d,CLP,0.2,9.2,17.2,16.8
chr18b,CLP/Gurobi,0.0,16.2,48.6,0.0,sko100b,CLP,0.0,8.5,17.7,17.9
chr15c,Gurobi,4.6,87.5,361.8,0.0,sko100f,CLP,0.2,8.2,16.6,16.9
chr15a,CLP/Gurobi,0.0,31.6,295.2,0.0,sko100a,CLP,0.1,8.0,18.4,15.9
chr15b,CLP/Gurobi,0.0,8.1,497.6,0.0,sko90,CLP,0.1,7.8,17.0,17.8
chr12b,CLP/Gurobi,0.0,28.6,255.4,0.0,sko72,CLP,0.1,8.6,20.8,8.7
chr12a,CLP/Gurobi,0.0,17.8,232.3,0.0,sko64,CLP,0.0,8.9,23.0,8.1
chr12c,CLP/Gurobi,0.0,82.8,103.0,0.0,sko56,CLP,0.1,8.9,23.2,7.2
els19,CLP/Gurobi,0.0,45.1,69.9,0.0,sko49,CLP,0.1,8.5,21.9,5.8
esc128,CLP,6.2,125.0,209.4,406.2,sko42,CLP,0.0,6.4,27.0,7.6
esc32e,CLP/SA/Gurobi,0.0,0.0,1400.0,0.0,ste36b,CLP,11.6,58.9,241.0,412.1
esc32g,CLP/SA/Gurobi,0.0,0.0,366.7,0.0,ste36c,CLP,4.2,27.0,67.6,100.4
esc16a,CLP/SA/Gurobi,0.0,0.0,38.2,0.0,ste36a,CLP,4.3,23.3,59.9,120.8
esc16j,CLP/SA/Gurobi,0.0,0.0,150.0,0.0,tai256c,CLP,0.3,5.8,120.5,16.9
had20,CLP,0.0,3.1,9.2,0.2,tai150b,CLP,0.8,14.8,30.2,30.2
had18,CLP,0.0,2.4,8.3,0.0,tai100b,CLP,0.2,19.1,51.1,37.5
had14,CLP/Gurobi,0.0,0.2,12.2,0.0,tai100a,CLP,1.1,8.5,13.3,14.7
had12,CLP,0.0,1.6,11.0,0.5,tai80b,CLP,0.2,18.8,53.2,44.0
kra32,CLP,4.7,15.9,42.0,54.0,tai80a,CLP,1.4,9.2,15.3,16.3
kra30b,CLP,1.1,10.4,37.7,10.3,tai64c,Gurobi,1.2,11.5,217.6,0.4
kra30a,CLP,2.9,11.8,38.0,12.6,tai60a,CLP,1.3,8.0,18.3,6.3
lipa90a,CLP,0.8,1.2,1.8,1.9,tai60b,CLP,0.0,19.7,65.1,40.4
lipa90b,CLP,0.0,25.3,30.8,30.7,tai50b,CLP,0.1,12.7,64.9,7.5
lipa80a,CLP,0.8,1.3,2.0,1.9,tai50a,CLP,1.3,9.4,17.7,6.8
lipa80b,CLP,0.0,25.2,31.3,30.2,tai40a,CLP,1.0,7.1,18.2,6.2
lipa70a,CLP,0.9,1.5,2.2,2.5,tai40b,CLP,0.0,16.7,79.7,8.2
lipa70b,CLP,0.0,24.4,29.6,30.2,tai35b,CLP,0.1,12.8,66.2,3.0
lipa60b,CLP/Gurobi,0.0,23.4,28.6,0.0,tai35a,CLP,0.7,7.7,17.9,5.6
lipa60a,CLP,1.1,1.5,2.6,1.4,tai30b,CLP,0.2,34.2,94.0,30.9
lipa50a,CLP,1.2,1.9,2.8,1.4,tai30a,CLP,1.1,7.3,18.9,6.7
lipa50b,CLP/Gurobi,0.0,21.8,27.9,0.0,tai25b,CLP,0.0,26.0,144.7,9.7
lipa40a,CLP,1.4,1.8,3.2,1.7,tai25a,CLP,0.9,6.4,19.1,5.5
lipa40b,CLP/Gurobi,0.0,21.5,29.3,0.0,tai20a,CLP,0.3,7.1,22.3,5.2
lipa30a,CLP,1.5,2.6,4.2,2.6,tai20b,CLP,0.0,26.4,186.5,1.0
lipa30b,CLP/Gurobi,0.0,19.1,24.2,0.0,tai17a,CLP,0.0,4.7,19.3,3.4
lipa20b,CLP/Gurobi,0.0,15.8,27.4,0.0,tai15a,CLP,0.0,6.2,19.7,3.8
lipa20a,CLP,0.0,3.1,6.7,3.1,tai15b,CLP,0.0,0.6,571.3,0.3
nug30,CLP,1.5,8.9,29.5,9.1,tai12b,CLP/Gurobi,0.0,5.9,60.9,0.0
nug28,CLP,0.1,7.9,30.8,8.2,tai12a,CLP/Gurobi,0.0,6.9,27.3,0.0
nug27,CLP,2.1,10.5,37.5,5.5,tho150,CLP,0.3,11.7,20.7,19.6
nug25,CLP,0.1,4.8,28.7,2.4,tho40,CLP,0.0,12.3,38.3,6.4
nug24,CLP,0.0,7.8,28.6,1.8,tho30,CLP,0.0,7.7,36.4,12.3
nug22,CLP,0.0,4.3,30.4,0.5,wil100,CLP,0.0,4.5,9.7,9.7
nug20,CLP,0.0,5.0,23.5,5.1,wil50,CLP,0.0,5.6,14.0,5.2
\end{filecontents*}
\csvreader[
    late after line=\\,
    separator=comma
  ]{QAPLIB_results.csv}{}
  {\csvcoli & \csvcolii & \csvcoliii & \csvcoliv & \csvcolv & \csvcolvi &
   \csvcolvii & \csvcolviii & \csvcolix & \csvcolx & \csvcolxi & \csvcolxii}

\end{longtable}
\normalsize


\subsection{Computational Cost Based Algorithm Generation}\label{secs26}
We have developed a modular, multi-node, multi-GPU computational framework for discovering novel algorithms utilizing ensemble MCTS, RL, and tokenization techniques. Our framework includes a profiling module designed to estimate the computational cost of individual primitives. By leveraging this profiling, we set a fixed computational budget for training an ensemble MCTS-based RL agent, enabling it to identify optimal sequences of primitives that minimize the optimality gap while adhering to computational constraints.
We compared our newly trained model against our previous approaches, including Reinforcement Learning with multi-layer perceptron (RL MLP), Simulated Annealing (SA), and branch and bound (BB). The new model significantly outperforms prior methods, achieving optimality gaps roughly two times smaller than state-of-the-art methods.
Supply chain optimization poses significant computational challenges, characterized as a strongly NP-hard combinatorial optimization problem, with complexity scaling factorially ($\mathcal{O}(n!)$). For networks exceeding 60 nodes, the solution space surpasses the number of atoms in the universe. Current state-of-the-art methods, such as simulated annealing (complexity $\mathcal{O}(n^4)$) and branch and bound (complexity $\mathcal{O}(2^n)$ in the worst case), become computationally prohibitive as problem size grows.
In contrast, our CLP approach automatically generates tailored algorithms with substantially better computational complexity, typically ranging from $\mathcal{O}(n^2)$ to $\mathcal{O}(n^3)$. This improvement corresponds to an average 10,000-fold reduction in computational effort (measured in FLOPS) for problems of size 100.
Additionally, the CLP method consistently achieves superior solution quality, frequently finding optimal solutions that state-of-the-art methods fail to identify. For instance, for problems of size 80, our approach reduces the optimality gap by a factor of approximately 10. Empirical evidence suggests that solution quality improvement scales quadratically with problem size, indicating an anticipated thousand-fold quality improvement for problems of size 250. Such advances directly translate into more efficient routing, optimized inventory allocation, reduced operational costs, and fewer human casualties during critical operational scenarios, including wartime logistics.
Extending our framework to quantum computing, we achieved significant improvements in quantum algorithms such as Grover’s algorithm, reducing the required number of qubits by a factor of two. This reduction is critical because quantum errors due to decoherence and noise scale exponentially with the number of qubits, thus dramatically enhancing algorithm robustness and practical reliability.

\end{document}